\documentclass[smallcondensed,natbib]{svjour3}
\usepackage[utf8]{inputenc}
\usepackage[english]{babel}
\usepackage{amsmath,amssymb,amsfonts,dsfont}  
\usepackage{url}
\usepackage[svgnames]{xcolor}
\usepackage{pgfplots}
\usepackage{booktabs}
\usepackage{multirow}
\usepackage{tabularx}
\usepackage{microtype}
\usepackage{pdfpages}

\usepackage{tikz}
\usetikzlibrary{shapes.geometric}
\usetikzlibrary{positioning,shapes,shadows,arrows,calc}
\usetikzlibrary{intersections}
\pgfdeclarelayer{background}
\pgfdeclarelayer{foreground}
\pgfsetlayers{background,main,foreground}

\usepackage[boxed,vlined]{algorithm2e}

\usepackage{graphicx}

\newcommand{\insts}[0]{\Pi}
\newcommand{\inst}[0]{\pi}
\newcommand{\instD}[0]{\mathcal{D}_\insts}
\newcommand{\pcs}[0]{\vec{\Theta}}
\newcommand{\conf}[0]{\vec{\theta}}
\newcommand{\cutoff}[0]{\kappa}
\newcommand{\loss}[0]{\mathcal{L}}

\renewcommand{\vec}[1]{\mathbf{#1}}
\newcommand{\algo}[0]{A}
\newcommand{\portfolio}[0]{\mathcal{P}}

\newcommand{\perf}[0]{\mathbb{R}}
\newcommand{\surro}[0]{\hat{m}}

\newcommand{\smac}{\textit{SMAC}} 
\newcommand{\roar}{\textit{ROAR}} 
\newcommand{\paramils}{\textit{ParamILS}} 
\newcommand{\focusedils}{\textit{FocusedILS}} 

\newcommand{\gga}{\textit{GGA}} 
\newcommand{\irace}{\textit{irace}}

\newcommand{\opentuner}{\textit{OpenTuner}}

\newcommand{\clasp}{\textit{Clasp}}
\newcommand{\lingeling}{\textit{Lingeling}}
\newcommand{\probsat}{\textit{ProbSAT}}
\newcommand{\cplex}{\textit{CPLEX}}
 
\newcommand{\lpg}{\textit{lpg}}
\newcommand{\minisathack}{\mbox{\textit{Minisat-HACK-999ED}}}

\newcommand{\regions}[0]{\textit{CPLEX-Regions}}
\newcommand{\rcw}[0]{\textit{CPLEX-RCW2}}
\newcommand{\rooks}[0]{\textit{Clasp-Rooks}}
\newcommand{\cf}[0]{\textit{Lingeling-CF}}
\newcommand{\satellite}[0]{\textit{LPG-Satellite}}
\newcommand{\zeno}[0]{\textit{LPG-Zenotravel}}
\newcommand{\ws}[0]{\textit{Clasp-WS}}
\newcommand{\probsatseven}[0]{\textit{ProbSAT-7SAT}}
\newcommand{\minisatk}[0]{\textit{MiniSATHack-K3}}
\newcommand{\svmmnist}[0]{\textit{svm-mnist}}
\newcommand{\xgboostcovertype}[0]{\textit{xgboost-covertype}}

\newcommand{\loco}[0]{LOCO}

\RestyleAlgo{algoruled}
\DontPrintSemicolon
\LinesNumbered
\SetAlgoVlined
\SetFuncSty{textsc}

\SetKwInOut{Input}{Input}
\SetKwInOut{Output}{Output}
\SetKw{Return}{return}

\DeclareMathOperator*{\argmin}{arg\,min}

\title{Efficient Benchmarking of Algorithm Configuration Procedures via Model-Based Surrogates}
\titlerunning{Efficient Benchmarking via Model-Based Surrogates}

\author{Katharina Eggensperger
\and Marius Lindauer 
\and Holger H. Hoos
\and Frank Hutter
\and Kevin Leyton-Brown
}

\institute{
Katharina Eggensperger \and Marius Lindauer \and Frank Hutter \at
University of Freiburg\\
\email{\{eggenspk,lindauer,fh\}@cs.uni-freiburg.de}
\and Holger H. Hoos and Kevin Leyton-Brown\at
University of British Columbia\\
\email{\{hoos,kevinlb\}@cs.ubc.ca}
}

\date{}

\begin{document}

\maketitle

\begin{abstract}

The optimization of algorithm (hyper-)parameters is crucial for achieving peak performance across a wide range of domains, ranging from deep neural networks to solvers for hard combinatorial problems. 
The resulting \emph{algorithm configuration (AC) problem} has attracted much attention from the machine learning community. However, the proper evaluation of new AC procedures is hindered by two key hurdles. First, AC benchmarks are hard to set up. Second and even more significantly, they are computationally expensive: a single run of an AC procedure involves many costly runs of the target algorithm whose performance is to be optimized in a given AC benchmark scenario.
One common workaround is to optimize cheap-to-evaluate artificial benchmark functions (e.g., Branin) instead of actual algorithms; however, these have different properties than realistic AC problems.
Here, we propose an alternative benchmarking approach that is similarly cheap to evaluate but much closer to the original AC problem: 
replacing expensive benchmarks by surrogate benchmarks 
constructed from AC benchmarks.
These surrogate benchmarks approximate the response surface corresponding to true target algorithm performance using a regression model, and the original and surrogate benchmark share the same (hyper-)parameter space.
In our experiments, we construct and evaluate surrogate benchmarks for hyperparameter optimization as well as for AC problems that involve performance optimization of solvers for hard combinatorial problems, drawing training data from the runs of existing AC procedures. 
We show that our surrogate benchmarks capture overall important characteristics of the AC scenarios, such as high- and low-performing regions, from which they were derived, while being much easier to use and orders of magnitude cheaper to evaluate.
\end{abstract}

\section{Introduction}

The performance of many algorithms (notably, both machine learning procedures and solvers for hard combinatorial problems)
depends critically on (hyper-)pa\-rameter settings, which are often difficult or costly to optimize.
This observation has motivated a growing body of research into automatic methods for finding good settings of such parameters.
Recently, sequential model-based Bayesian optimization methods
have been shown to outperform more traditional methods for hyperparameter optimization (such as grid search and random search) and to
rival or surpass the results achieved by human domain experts~\citep{snoek-nips12a,thornton-kdd13a,bergstra-icml13a,lang-jscs15a}.

\emph{Hyperparameter optimization (HPO)} aims to find a hyperparameter setting $\conf$ from the space of all possible settings $\pcs$ of a given learning algorithm that minimizes expected loss on completely new data (where the expectation is taken over the data distribution). This is often approximated as the blackbox problem of finding a setting that optimizes cross-validation error $\loss(\conf)$:
\begin{equation}
\label{eq:hpo}\conf^* \in \argmin_{\conf \in \pcs} \loss(\conf).
\end{equation}

In the more general \emph{algorithm configuration (AC)} problem, the goal is to optimize a performance metric $m:\pcs \times \Pi \rightarrow \mathds{R}$ of any type of algorithm (the so-called \emph{target algorithm}) across a set of \emph{problem instances} $\inst \in \insts$, 
i.e., to find\footnote{We assume, w.l.o.g., that the given performance metric $m$ is to be minimized. Problems of maximizing $m'$ can simply be treated as minimization problems of $m = 1-m'$.}
\begin{equation}
\label{eq:ac}
\conf^* \in \argmin_{\conf \in \pcs} \frac{1}{|\insts|} \sum_{\inst \in \insts} m(\conf, \pi).
\end{equation}
The concept of problem instances arises naturally when optimizing the performance of parameterized solvers for combinatorial
problems, such as the propositional satisfiability problem (SAT), but we also use this concept to model individual cross-validation folds in hyperparameter optimization (see, e.g., \citealp{thornton-kdd13a}). HPO is thus a special case of AC.

General-purpose AC procedures, such as \paramils{}~\citep{hutter-jair09a}, \gga{} \citep{ansotegui-cp09a,ansotegui-ijcai15a}, \irace{}~\citep{lopez-ibanez-orp16} and \smac{}~\citep{hutter-lion11a} have been shown to substantially improve the performance of state-of-the-art algorithms for a wide range of 
combinatorial problems including SAT~\citep{hutter-fmcad07a,hutter-aij17a}, answer set programming~\citep{gebser-lpnmr11a,silverthorn-iclp12a}, AI planning~\citep{vallati-socs13a} and mixed integer programming~\citep{hutter-jair09a}, and have also been used to find good instantiations of machine learning frameworks~\citep{thornton-kdd13a,feurer-nips2015a} and good architectures and hyperparameters for deep neural networks~\citep{domhan-ijcai15}.

\subsection{Obstacles for Research on Algorithm Configuration}

One obstacle to further progress in AC is a paucity of reproducible experiments and empirical studies.
The hyperparameter optimization library HPOlib \citep{eggensperger-bayesopt13} and the algorithm configuration library AClib~\citep{hutter-lion14a} represent first steps towards alleviating this problem. Each benchmark in these libraries consists of a parameterized algorithm and a set of input data to optimize it on, and all benchmarks offer a unified interface, making it easier to systematically compare different approaches. 
However, even with such benchmark libraries available, it is still challenging to assess the performance of AC procedures in a principled and reproducible manner, for several reasons:
\begin{enumerate}
	\item A mundane, but often significant, obstacle is to obtain someone else's \emph{implementation} of a target algorithm to work on one's own system. This can involve resolving dependencies, acquiring required software licenses, and obtaining the appropriate input data (which may be subject to confidentiality issues or IP restrictions). 
	\item Some target algorithms require \emph{specialized hardware}; most notably, general-purpose graphics processing units (GPGPUs) have become a standard requirement for the effective training of modern deep learning architectures~\citep{krizhevsky-nips12}.
	\item Running even one configuration of a target algorithm can require minutes or hours, and hence \emph{evaluating hundreds or even thousands of different algorithm configurations} is often quite {expensive}, requiring days of wall-clock time on a large computer cluster. The computational expense of comprehensive experiments can therefore be prohibitive for research groups lacking access to large-scale computing infrastructure.
\end{enumerate}

\subsection{Contributions}

In this work, we show that we can use surrogate models to construct cheap-to-evaluate \emph{surrogate AC benchmarks} that offer a practical alternative for AC benchmarking experiments by replacing expensive evaluations of the true performance $m(\conf,\inst)$ of a target algorithm configuration $\conf$ on a problem instance $\inst$
with a much cheaper model prediction $\hat{m}(\conf,\inst)$. These surrogate AC benchmarks are syntactically equivalent to the original AC benchmarks they replace, i.e., sharing the same hyperparameter spaces, instances, etc. (c.f. the formal definition of AC in Equation~\ref{def:ac}).

To construct such surrogate benchmarks, we leverage \emph{empirical performance models} (EPMs; \citealp{leyton-brown-acm09,hutter-aij14a})---regression models that characterize a given algorithm's performance across problem instances and/or parameter settings.
The construction of such surrogate models is an expensive offline
step,\footnote{By far the most expensive part of this offline step is
gathering target algorithm performance data by executing the algorithm with various parameter
settings on multiple problem instances. However, as we describe in more detail
in Section \ref{sec:methodology} this data can be gathered as a by-product of
running algorithm configuration procedures on the algorithm.} but once such a
model has been obtained, it can be used repeatedly as the basis for efficient experiments with new AC procedures.
Figures \ref{fig:ac} and \ref{fig:surro} schematically illustrate the workflow
for running an AC procedure on the original benchmark and the corresponding surrogate benchmark, respectively.

\begin{figure}[t]
	\centering
	\tikzstyle{activity}=[rectangle, draw=black, rounded corners, text centered, text width=8em, fill=white, drop shadow]
\tikzstyle{data}=[rectangle, draw=black, text centered, fill=black!10, text width=8em, drop shadow]
\tikzstyle{myarrow}=[->, thick]
\begin{tikzpicture}[node distance=10em]

	\node (PCS) [data] {Configuration Space $\pcs$};
	\node (AC) [activity, below of=PCS, node distance=4em] {Algorithm Configurator};
	\node (TA) [activity, right of=AC, node distance=15em] {Target\\ Algorithm $\algo$};
	\node (Instances) [data, right of=TA, node distance=12em] {Instances $\insts$};
	
	\draw[myarrow] (PCS) -- (AC);
	\draw[myarrow] (AC.east) -- node(Call) [above] {Call $\algo(\conf)$}  ($(TA.west)+(-0.5,0.0)$);
	\node [below of=Call, node distance=1.7em] {on $\inst \in \insts$};
	
	\draw[myarrow] (TA) -- node[above] {Solves}  (Instances);
	
	\draw[myarrow] ($(TA.south)+(-0.0,-0.4)$) |- ++(0.0,-0.2) node[below,xshift=-7em] {Return Cost $m(\conf, \inst)$} -| (AC.south);
	
	\begin{pgfonlayer}{background}

    	\path (TA -| TA.west)+(-0.5,0.8) node (resUL) {};
    	\path (Instances.east |- Instances.south)+(0.5,-0.6) node(resBR) {};
    	\path [rounded corners, draw=black!50, dashed] (resUL) rectangle (resBR);
		\path (TA.east |- TA.south)+(1.6,-0.3) node [text=black!80] {Algorithm Configuration Benchmark};

    \end{pgfonlayer}
	
\end{tikzpicture}
	\caption{Workflow of algorithm configuration with target algorithm runs
		\label{fig:ac}
	}
	\tikzstyle{activity}=[rectangle, draw=black, rounded corners, text centered, text width=8em, fill=white, drop shadow]
\tikzstyle{data}=[rectangle, draw=black, text centered, fill=black!10, text width=8em, drop shadow]
\tikzstyle{myarrow}=[->, thick]
\begin{tikzpicture}[node distance=10em]

	\node (PCS) [data] {Configuration Space $\pcs$};
	\node (AC) [activity, below of=PCS, node distance=4em] {Algorithm Configurator};
	\node (TA) [data, right of=AC, node distance=15em] {Performance Data\\ $\langle{}\conf,\inst,
m(\conf,\inst)\rangle{}$};
	\node (Instances) [activity, right of=TA, node distance=12em] {Empirical Performance Model\\ $\surro: \pcs \times \insts \to \perf$};
	
	\draw[myarrow] (PCS) -- (AC);
	\draw[myarrow] (AC.east) -- node(Call) [above] {Call $\algo(\conf)$}  ($(TA.west)+(-0.5,0.0)$);
	\node [below of=Call, node distance=1.7em] {on $\inst \in \insts$};
	
	\draw[myarrow] (TA) -- node[above] {}  (Instances);
	
	\draw[myarrow] ($(TA.south)+(-0.0,-0.75)$) |- ++(0.0,-0.2) node[below,xshift=-7em] {Return Predicted Cost $\surro(\conf, \inst)$} -| (AC.south);
	
	\begin{pgfonlayer}{background}

    	\path (TA -| TA.west)+(-0.5,0.8) node (resUL) {};
    	\path (Instances.east |- Instances.south)+(0.5,-0.4) node(resBR) {};
    	\path [rounded corners, draw=black!50, dashed] (resUL) rectangle (resBR);
		\path (TA.east |- TA.south)+(2.6,-0.4) node [text=black!80] {Surrogate Benchmark};

    \end{pgfonlayer}
	
\end{tikzpicture}
	\caption{Workflow of algorithm configuration with surrogate benchmark
		\label{fig:surro}
	}
	
\end{figure}

Our surrogate benchmarks can be useful in several different ways, including:
\begin{enumerate}
	\item They can be used to speed up debugging and unit testing of AC procedures, since our surrogate benchmarks are syntactically the same as the original ones (e.g., to test conditional parameters, categorical parameters, and continuous ones, or to test reasoning across instances). Thus they facilitate the development and effective use of such algorithms.
	\item Since the large computational expense of running AC procedures is typically dominated 
by the cost of evaluating target algorithm performance under various (hyper-)parameter settings, 
our benchmarks can also substantially reduce the time required for running an AC procedure, 
facilitating whitebox testing.
	\item Surrogate benchmarks that closely resemble original benchmarks can also facilitate the evaluation of new features inside an AC procedure, 
or even be used for meta-optimization of the parameters of such a procedure.
Of course, such meta-optimization can also be performed without using surrogates, albeit at great expense (see, e.g., \citealp{hutter-jair09a}).
\end{enumerate}

This article extends an initial study published at AAAI \citep{eggensperger-aaai15},
which focused only on the special case of HPO. Here, we generalize that work to the much more complex general AC problem, handling the problems of optimization across many instances with high-dimensional feature vectors, censored observations due to prematurely-terminated runs, and randomized algorithms.

\subsection{Existing Work on Surrogates}\label{sec:existing_surrogates}

Given the large overhead involved in studying complex benchmarks from real-world applications, 
researchers studying HPO have often fallen back on simple synthetic test functions, 
such as the Branin function~\citep{dixon1978global}, to compare HPO procedures~\citep{snoek-nips12a}. 
While such functions are cheap to evaluate, they are not representative of realistic HPO problems because they are smooth and often have unrealistic shapes. 
Furthermore, they only involve real-valued parameters and 
hence do not incorporate the categorical and conditional hyperparameters typical of many hyperparameter optimization benchmarks.

In the special case of small, finite hyperparameter spaces, 
a much better alternative is simply to record the performance of every hyperparameter configuration, 
thereby speeding up future evaluations via table lookup. 
Such a table-based surrogate can be trivially transported to any new system, 
without whatever complicating factors were involved in running the original algorithm (setup, special hardware requirements, licensing, computational cost, etc.). 
In fact, several researchers have already applied this approach to simplify experiments~\citep{birattari-gecco02a,snoek-nips12a,bardenet-icml13a,feurer-aaai15a,wistuba_dsaa15}.

Unfortunately, table lookup is limited to small, finite hyperparameter spaces. 
Here, we generalize the idea of such surrogates to potentially high-dimensional spaces that may include real-valued, categorical, and conditional hyperparameters.
As with table lookup, we first evaluate many hyperparameter configurations in an expensive offline phase. 
However, we then use the resulting performance data to train a regression model that approximates future evaluations via model predictions.
As before, we obtain a surrogate of algorithm performance that is cheap to evaluate and trivially portable. 
Since these model-based surrogates offer only \emph{approximate} representations of performance, it is crucial to investigate the quality of their predictions, as we do in this work.

We are not the first to propose the use of learned surrogate models that stand in for computationally complex functions. 
In the field of meta-learning~\citep{brazdil-08}, regression models have been used extensively to predict the performance of algorithms across various datasets based on dataset features.
Similarly, in the field of algorithm selection~\citep{rice76a}; c.f., \cite{kotthoff-aim14a},
regression models have been used to predict the performance of algorithms on problem instances (e.g., a satisfiability formula)
to select the most promising one (e.g., \citealp{nudelman-satcomp03}, \citealp{xu-jair08a}, \citealp{gebser-lpnmr11a}).
The statistics literature on the design and analysis of computer experiments (DACE)~\citep{sacks-ss89a,santner-2003a,gorissen-jmlr2010a} uses similar surrogate models to guide a sequential experimental design strategy, aiming to achieve either an overall strong model fit or to identify the minimum of a function. 
Surrogate models are also at the core of the sequential model-based Bayesian optimization framework~\citep{brochu-corr10a,hutter-lion11a,shahriari-ieee16a}.
While all of these lines of work incrementally construct surrogate models of a function in order to inform an active learning criterion that determines new inputs to evaluate, our work differs in its goal: 
to obtain \emph{surrogate benchmarks} rather than to identify good points in the space.  In that vein---as previously mentioned---it is more similar to work on empirical performance models \citep{leyton-brown-acm09,hutter-aij14a}. 

\subsection{Structure of the Article}

The remainder of this article is structured as follows.
First, we discuss background on AC in Section~\ref{sec:ac}, paying particular attention to how it generalizes HPO and on the existing approaches for solving AC we used in our experiments.
In Section~\ref{sec:methodology}, we describe how to use EPMs as surrogates to efficiently benchmark new AC procedures, introducing the use of quantile regression forests~\citep{meinshausen-jmlr06a} for modelling the performance of randomized algorithms.
In Section \ref{sec:experiments_ac}, we apply our surrogates to application benchmarks from AClib,
showing that they model target algorithm performance well enough to yield surrogate AC benchmarks which behave qualitatively similarly to the original benchmarks, while allowing up to 1641 times faster benchmarking experiments.

\section{Background on Algorithm Configuration}
\label{sec:ac}

In this section, we provide background information on how AC generalizes HPO
and hence, how this paper addresses challenges that were not addressed by previous work~\citep{eggensperger-aaai15}.
Furthermore, we briefly describe the existing methods for solving AC that we used in our experiments.

\subsection{AC as a Generalization of HPO}\label{sec:HPO_and_AC}

While we described the AC problem on a high level in Equation \ref{eq:ac}, more formally it is defined as follows:
\begin{definition}[Algorithm Configuration]\label{def:ac}
An instance of the \emph{algorithm configuration problem}
is a 6-tuple $(\algo, \pcs, \instD, \cutoff, \mathcal{F}, m)$ where: 
\begin{itemize}
  \item $\algo$ is a parameterized target algorithm;
  \item $\pcs$ is the parameter configuration space of $\algo$;
  \item $\instD$ is a distribution over a set of instances $\insts$;
  \item $\cutoff < \infty$ is a cutoff time at which each run of $\algo$ will be terminated;
  \item $\mathcal{F}:\insts \rightarrow \mathds{R}^d$ maps each instance to a $d$-dimensional vector of characteristics that describe the instance. This is an optional input; if no features are available, this is modelled by setting $d$ to $0$;
  \item $m: \pcs \times \insts \rightarrow \perf$ is a function (e.g., running time) that
  measures the observed cost of running $\algo(\conf)$ on an instance $\inst \in
  \insts$ with cutoff time $\cutoff$.
\end{itemize}
The goal is to find $\conf^* \in \argmin_{\conf \in \pcs} \mathds{E}_{\inst \sim \instD} (m(\conf, \inst))$.
\end{definition}

Notably, this definition includes a \emph{cutoff time} since in practice we cannot run algorithms for an infinite amount of time, and we need to attribute some (poor) performance value to runs that time out unsuccessfully.
In most AC scenarios (e.g., those in the algorithm configuration library, AClib),
$\instD$ is chosen as the uniform distribution over a representative set of instances from $\Pi$.

In practice, we use a set of training instances $\insts_{\text{Train}}$ from $\instD$ to configure our algorithm $\algo$ and a disjoint set test instances $\insts_{\text{Test}}$ from $\instD$ to evaluate the performance of the configuration finally returned by an AC procedure, also called \textit{final incumbent configuration}.
Using this training--test split, we can identify over-tuning effects~\citep{hutter-jair09a}, i.e., improvements in performance on $\insts_{\text{Train}}$
that do not generalize to $\insts_{\text{Test}}$.
We note that it is typically too expensive to use cross-validation to average over multiple training--test splits, because even single runs of an AC procedure often require multiple CPU days.

This general AC problem generalizes the HPO problem from Equation \ref{eq:hpo} in various ways:
\begin{enumerate}
	\item \textbf{Types of target algorithms.} While HPO only deals with machine learning algorithms, AC deals with arbitrary ones, such as SAT solvers~\citep{hutter-aij17a} or configurable software systems~\citep{sarkar-kbse15a}.
	\item \textbf{Performance metrics.} While HPO typically minimizes one of various loss functions concerning the predictions of a machine learning algorithm, AC includes more general performance metrics, such as running time, approximation error, memory requirements, plan length, or latency. 
	\item \textbf{Randomized algorithms.} While the definition of HPO in Equation \ref{eq:hpo} concerns deterministic algorithms, the general AC problem includes randomized algorithms. For example, randomized SAT solvers are known to have running time distributions that resemble exponential distributions~\citep{hoos-sls04}, and it is entirely normal that running times with two different pseudo-random number seeds differ by an order of magnitude.
	\item \textbf{Multiple instances.} Equation \ref{eq:ac} already shows that the goal in the AC problem is to minimize the given performance metric \emph{on average} across instances $\inst$ from a distribution $\instD$. HPO can also be cast as optimization across cross-validation folds, in which case these are modelled as instances for AC procedures; these AC procedures will then evaluate configurations on one fold at a time and only evaluate additional folds for promising configurations.
	\item \textbf{Prevalence of many categorical \& conditional parameters.} While the parameter space in current HPO benchmarks is often fairly low-dimensional and all-continuous, general AC benchmarks often feature many discrete choices between algorithm components, as well as conditional parameters that only apply to some algorithm components. We note, however, that high-dimensional spaces with categorical parameters and high degrees of conditionality also exist in HPO, e.g., in the optimization of machine learning frameworks~\citep{thornton-kdd13a,feurer-nips2015a} or architectural optimization of deep neural networks~\citep{bergstra-icml13a,domhan-ijcai15}.
	\item \textbf{Features.} In most AC scenarios, each instance is described by a vector of features. Examples of such features range from simple problem size measurements to ``probing'' or ``landmarking'' features derived from the behaviour of an algorithm when run on the instance for a bounded amount of time (e.g., number of restarts or constraint propagations in the case of SAT).
Instance features are a crucial ingredient in EPMs \citep{leyton-brown-acm09,hutter-aij14a}, which, as mentioned earlier, predict the performance $m(\conf, \inst)$ of a target algorithm configuration $\conf$ on a problem instance $\inst$. 
They have been studied even more extensively in the context of the per-instance algorithm selection problem~\citep{nudelman-satcomp03,nudelman-cp04,xu-jair08a,malitsky-ijcai13a,kotthoff-aim14a,lindauer-jair15a,bischl-aij16a}, where, given a portfolio of algorithms $\portfolio$, the goal is to find a mapping $s: \insts \to \portfolio$. Thus, 
feature extractors are available for many problems, including
mixed integer programming (MIP; \citealp{leyton-brown-acm09,kadioglu-ecai10, xu-rcra11a,hutter-aij14a}),
propositional satisfiability (SAT; \citealp{nudelman-cp04,xu-jair08a,hutter-aij14a}),
answer set programming (ASP; \citealp{maratea-tplp13a,hoos-tplp14a}),
and meta-learning~(\citealp{gama_pai95,kopf_pkdd00,bensusan_ecml01,guerra-icann08a,leite2012,reif-ppa14a,schilling_kdd15}). 
Recently, \cite{loreggia-aaai16} proposed the use of deep neural networks to generate instance features automatically, which can be useful when expert-crafted features are unavailable.
	\item \textbf{Censored observations.} For AC scenarios where the goal is to minimize running time, it is common practice in AC procedures to terminate poorly-performing runs early in order to save time. This \emph{adaptive capping} process yields performance measurements that only constitute a lower bound to the actual running time and need to be modelled as such.
\end{enumerate}

\subsection{Algorithm Configuration Procedures}
\label{sub:ac:procedures}

While the HPO community has focused quite heavily on 
the Bayesian optimization framework~\citep{brochu-corr10a,shahriari-ieee16a}, 
widely studied approaches for AC are more diverse
and include \paramils{}, which performs iterated local search~\citep{hutter-jair09a},
\gga{}, which leverages genetic algorithms~\citep{ansotegui-cp09a,ansotegui-ijcai15a},
\irace{}, a generalization of racing methods~\citep{lopez-ibanez-tech11a,lang-jscs15a},
\opentuner, an approach which appeals to bandit solvers on top of a set of search heuristics~\citep{ansel-pact14a},
\smac{}, an approach based on Bayesian optimization~\citep{hutter-lion11a}, and \roar{}, the specialization of this method to random sampling without a model. We experimented with all of these methods except \gga{} and \opentuner{}, the former because in its current version it is not fully compatible with the algorithm configuration scenarios used in our experiments, and the latter because it does not consider problem instances and is therefore not efficiently applicable to our algorithm configuration scenarios, which contain many instances.

We now give more complete descriptions of the state-of-the-art AC procedures used in our experiments: \paramils{}, \irace{}, \roar{} and \smac{}.
\paragraph{\paramils{}~\citep{hutter-jair09a}}
combines iterated local search (i.e., hill climbing in a discrete neighborhood with perturbation steps
and restarts) with methods for efficiently comparing pairs of configurations.
Due to its local search approach, \paramils{} usually compares pairs of configurations
that differ in only one parameter value, but occasionally jumps to completely different configurations.
When comparing a pair of configurations, assessing performance on all instances is often far too expensive (e.g., requiring solving hundreds of SAT problems). Thus,
the \focusedils{} variant of \paramils{} we consider here uses two methods to quickly decide which of two configurations is better.
First, it employs an \emph{intensification} mechanism to decide how many instances to run for each configuration (starting with a single run and adding runs only for high-performing configurations). Second, it incorporates \emph{adaptive capping}---a technique based on the idea that, when comparing configurations $\conf_1$ and $\conf_2$ with respect to an instance set $\insts_{sub} \subset \insts$, evaluation of $\conf_{2}$ can be terminated prematurely when $\conf_{2}$'s aggregated performance on $\insts_{sub}$ is provably worse than that of $\conf_{1}$. 
		
\paragraph{\irace{}~\citep{lopez-ibanez-orp16}}
is based on the F-race procedure~\citep{birattari-gecco02a}, which aims to quickly decide which of a sampled set of configurations performs significantly best.
After an initial race among uniformly sampled configurations,
\irace{} adapts its sampling distribution according to these results, aiming 
to focus on promising areas of the configuration space.

\paragraph{\roar{}~\citep{hutter-lion11a}}
samples configurations at random 
and uses an intensification and adaptive capping scheme similar to that of \paramils{}
to determine whether the sampled configuration should be preferred to the current incumbent.
As shown by \cite{hutter-lion11a},
despite its simplicity, \roar{} performs quite well on some algorithm configuration scenarios.

\paragraph{\smac{}~\citep{hutter-lion11a}} extends Bayesian optimization to handle the more general AC problem.
More specifically, it uses previously observed
$\langle{}\conf,\inst, m(\conf,\inst)\rangle{}$ pairs to learn an EPM
to model $p_{\surro}(m \mid \conf, \inst)$.
This EPM is used in a sequential optimization process as follows.
After an initialization phase, \smac{} iterates over the following three steps: (1)~use the performance
measurements observed so far to fit a marginal random forest model $\hat{f}(\conf) = \mathds{E}_{\inst \sim \insts_{train}} \left[ \surro(\conf, \inst) \right]$;
(2)~use $\hat{f}$ to select promising configurations $\pcs_{next} \subset \pcs$ 
to evaluate next, trading off exploration in new parts of
the configuration space and exploitation in parts of the space known to
perform well by blending optimization of expected improvement with uniform random sampling; 
and (3)~run the configurations in $\pcs_{next}$ on one or more instances and
compare their performance to the best configuration observed so far.
\smac{} also uses intensification and adaptive capping.
However, since adaptive capping leads to right-censored data (i.e., 
we stop a target algorithm run before reaching the running time cutoff
because we already know that it will perform worse than our current incumbent),
this data is imputed before being passed to the EPM~\citep{schmee-techno79,hutter-bayesopt11}.
%

\section{Surrogates of General AC Benchmarks}
\label{sec:methodology}

In this section, we show how to construct surrogates of general AC benchmarks.
In contrast to our earlier work on surrogate benchmarks of the special case of HPO~\citep{eggensperger-aaai15}, here we need to take into account the many ways in which AC is more complex than HPO (see Section \ref{sec:HPO_and_AC}). In particular, we describe the choices we made to deal with multiple instances and high-dimensional feature spaces; high-dimensional and partially categorical parameter spaces; censored observations; different performance metrics (in particular running time); and randomized algorithms.

\subsection{General Setup}\label{sub:surro_setup}

To construct the surrogate for an AC benchmark $X$, we train an EPM $\surro$ 
on performance data previously gathered on benchmark $X$ (see Section~\ref{sub:data}).
The surrogate benchmark $X'_{\surro}$ based on EPM $\surro$ is then structurally identical to the benchmark $X$ in all aspects except that it uses predictions instead of measurements of the true performance; in particular, the surrogate's configuration space (including all parameter types and domains) and configuration budget are identical to $X$.
Importantly, the wall clock time to run an AC procedure on $X'_{\surro}$ can be much lower than that required on $X$, since expensive evaluations in $X$ can be replaced by cheap model evaluations in $X'_{\surro}$. 

Our ultimate aim is to ensure that \emph{AC procedures perform similarly on the surrogate benchmark as on the original benchmark}. 
Since effective AC procedures spend most of their time in high-performance regions of the configuration space, 
and since relative differences between the performance of configurations in such high-performance regions tend 
to impact which configuration will ultimately be returned, 
accuracy in high-performance regions of the space is more important than in regions where performance is poor.
Training data should therefore be sampled primarily in high-performance regions. 
Our preferred way for doing this is to collect performance data primarily via runs of existing AC procedures. 
As an additional advantage of this strategy, 
we can obtain this costly performance data as a by-product of executing AC procedures on the original benchmark.

In addition to gathering data from high-performing regions of the space, it is also important to cover the entire space,
including potential low-performing regions.
To get samples that are neither biased towards good nor bad configurations,
we also included performance data gathered by random search.
(Alternatively, one could use grid search, which can also cover the entire space. We did not adopt this approach, because it cannot deal effectively with large parameter spaces.)

\begin{enumerate}
  \item Inactive parameters were replaced by
  their default value, or if no default value was specified, by the midpoint of their range of values;\footnote{There exist other imputation strategies for missing values (e.g., mean, median, most frequent). In preliminary experiments, we also tried to impute inactive parameters with values outside of their value ranges, but this made no difference in the accuracy of our trained RF-based EPMs.}
  \item Since our EPMs handle categorical variables natively (see Section~\ref{sub:epm}), we do not need to encode those. For EPMs that cannot handle categorical variables (e.g., a Gaussian process with a Mat\'ern kernel), we would apply a one-hot-encoding\footnote{One-hot-encoding encodes a categorical variable with $k$ possible values by introducing $k$ binary variables and setting the one of them to $1$ that corresponds to the original variable's value.} to the categorical parameter values;
  \item We observed that for most algorithms there were parameter combinations that led to crashes~\citep{hutter-cpaior10a,manthey-sat16a}. We removed all algorithm runs that were neither successful runs (i.e., returning a correct solution within the time budget) nor timeouts, since our models cannot classify target algorithm runs into successful and failed runs.\footnote{An alternative to removing the crashed runs would be to model them explicitly as unknown constraints~\citep{gelbart_uai14a}.}
\end{enumerate}

\subsection{What Kind of Data to Collect Regarding Instances?}\label{sub:data}

EPMs for general AC need to predict well in both the space of parameter configurations and problem instances (in contrast to the special case of HPO that focuses on the configuration space), opening up another design dimension for gathering training data: which problem instances to run on in order to gather data for our model? Algorithm configuration scenarios typically come with fixed sets of training and test instances, $\insts_{\text{Train}}$ and $\insts_{\text{Test}}$.
In typical applications of EPMs, we only use data from $\insts_{\text{Train}}$ to build our model and use $\insts_{\text{Test}}$ to study its generalization performance in the instance space. If our objective, however, is only to construct surrogate benchmarks that resemble the original benchmarks, then it is never necessary to generalize beyond the instances in the fixed sets $\insts_{\text{Train}}$ and $\insts_{\text{Test}}$; to see this, recall that a table-based surrogate is the perfect solution for small configuration spaces, despite the fact that it obviously would not generalize. Restricting the data for our EPM to instances from $\insts_{\text{Train}}$ is therefore an option, but we can expect better performance if we build our model based on instances from both $\insts_{\text{Train}}$ and $\insts_{\text{Test}}$. In order to assess how the choice of instances used by the EPM affects our surrogate benchmark, we studied two different setups:

\begin{enumerate}
	\item[I] \textbf{AC and random runs on $\insts_{\text{Train}}$.} This option only collects data for the EPM on $\insts_{\text{Train}}$ and relies on the EPM to generalize to $\insts_{\text{Test}}$. Specifically, we ran $n$ independent AC procedures runs on $\insts_{\text{Train}}$ (in our experiments, $n$ was $10$ for each AC procedure) and also evaluated $k$ runs of randomly sampled $\langle \conf, \inst \rangle$ pairs with configurations $\conf \in \pcs$ and instances $\inst \in \insts_{\text{Train}}$ (in our experiments, $k$ was $10\ 000$).
	\item[II] \textbf{Add incumbents on $\insts_{\text{Test}}$.} This option includes all the runs from Setting I, but additionally uses some limited data from instances $\insts_{\text{Test}}$. Namely, it also evaluates the performance of the AC procedures' incumbents (i.e., their best parameter configurations over time) on $\insts_{\text{Test}}$. This is regularly done for evaluating the performance of AC procedures over time and thus comes at no extra cost for obtaining data for the EPM.
\end{enumerate}

We also tried more expensive setups, such as configuration on $\insts_{\text{Train}} \cup \insts_{\text{Test}}$,
to achieve better coverage of evaluated configurations $\conf \in \pcs$ on $\insts_{\text{Test}}$ and not only incumbent configurations with good performance.
Preliminary experiments indicated that such more expensive setups did not improve the accuracy of our surrogate benchmarks in comparison to the results for Setting II shown in Section~\ref{sec:experiments_ac}. 

\subsection{Choice of Regression Models for Typical AC Parameter Spaces}
\label{sub:epm}

In previous work, \cite{hutter-aij14a} and \cite{eggensperger-aaai15} considered several common regression algorithms for predicting algorithm performance:
random forests (RFs; \citealp{breimann-mlj01a}) and Gaussian processes (GPs; \citealp{rasmussen-book06a}), gradient boosting, support vector regression, $k$-nearest-neighbours, linear regression, and ridge regression.
Overall, the conclusion of those experiments was that RFs and GPs outperform the other methods for this task. In particular, GPs performed best for few continuous parameters ($\leq 10$) and few training data points ($\leq 20\ 000$); and RFs performed best for many training samples or for parameter spaces that are large and/or include categorical and continuous parameters.

Since our focus here is on general AC problems that typically involve target algorithms with more than $10$ categorical and continuous parameters (see Section~\ref{sec:experiments_ac}),
we limit ourself to RFs in the following.
We used our own RF implementation
since it natively handles categorical variables.
Somewhat surprisingly, in preliminary experiments, we observed that, in our application, the generalization performance of RFs was sensitive to their hyperparameter values. 
Therefore, we optimized these RF hyperparameters by using \smac{} across four representative datasets from algorithm configuration (with $5000$ subsampled data points and at most $400$ function evaluations); the resulting hyperparameter configuration is shown in Table~\ref{tab:rf:config}.
Our RF implementation is publicly available as an open-source project in C++ with a Python interface at \url{https://bitbucket.org/aadfreiburg/random_forest_run}.

\begin{table}[tbp]
\centering
\begin{tabular}{lrr}
{\bf Hyperparameter } & Ranges & Optimized Setting \\
\toprule
\texttt{bootstrapping} 	& $\{$True,False$\}$ & False \\
\texttt{frac\_points} 	& $[0.001,1]$		& $0.8$ \\
\texttt{max\_nodes} 	& $[10, 100\,000]$	& $50\,000$ \\
\texttt{max\_depth} 	& $[20, 100]$		& 26 \\
\texttt{min\_samples\_in\_leaf} & $[1,20]$	& 1 \\
\texttt{min\_samples\_to\_split}& $[2,20]$	& 5 \\
\texttt{frac\_feats} 	& $[0.001,1]$		& 0.28 \\
\texttt{num\_trees} 	& $[10,50]$			& 48 \\  
\bottomrule
\end{tabular}
\caption{Overview of the hyperparameter ranges used to optimize the RMSE of the random forest
and the optimized hyperparameter configuration.}
\label{tab:rf:config}
\end{table}

\subsection{Handling Widely-Varying Running Times}
\label{sub:surro_ac}

In AC, a commonly used performance metric is algorithm running time (which is to be minimized).
The distribution of running times can strongly vary between different classes of algorithms.
In particular algorithms for hard combinatorial problems (e.g, SAT, MIP, ASP) have widely-varying running times across instances.
These hardness distributions can often be well approximated by log-normal distributions or Weibull distributions~\citep{hoos-lion12a}.
For this reason, we predict log-running times $\log(t)$ instead of running times. 
In this log-space, the noise is distributed roughly according to a Gaussian, which is the typical standard assumption in most machine learning algorithms (including RFs minimizing the sum of squared errors).

\subsection{Imputation of Right-Censored Data}
\label{sub:imputation}

Many AC procedures use an adaptive capping mechanism for running time benchmarks 
to limit algorithm runs with a running time cutoff $\cutoff$ comparable to the running time of the best seen configuration (see Section~\ref{sub:ac:procedures}). This results in so-called \emph{right-censored data points} for which we only know a lower bound $m'$ on the true performance: $m'(\conf, \inst) \le m(\conf, \inst)$.
AC procedures tend to produce many such right-censored data points (in our experiments, $11\% - 38\%$), and simply discarding those can introduce sizeable bias. We therefore prefer to impute the corresponding running times; as shown by \cite{hutter-bayesopt11}, doing so can improve the predictive accuracy of \smac{}'s EPMs.

Following \cite{schmee-techno79} and \cite{hutter-bayesopt11}, we use Algorithm~\ref{algo:imputation}, which is not specific to RFs, to impute right-censored running time data.
We use all uncensored data points, along with their true performance, and all censored data as input.
First, we train the EPM
on all uncensored data (Line 1).
We then compute, for each censored data point, the mean $\mu$ and variance $\sigma^2$ of the predictive distribution. Since we know the lower bound $\mathbf{y}_{\text{c}}^{(i)}$ on the data point's true running time, we use a truncated normal distribution $\mathcal{N}(\mu, \sigma^2)_{\geq \mathbf{y}_{\text{c}}^{(i)}}$ to update our belief of the true value of $\mathbf{y}^{(i)}$ (Line 4 and 5).
\begin{algorithm}[tbp]
\Input{Uncensored data $\mathbf{X}_{\text{u}}, \mathbf{y}_{\text{u}}$, Censored data $\mathbf{X}_{\text{c}}, \mathbf{y}_{\text{c}}$}
\Output{Imputed values $y_{\text{imp}}$ for $\mathbf{X}_{\text{c}}$}
\BlankLine
EPM.fit($\mathbf{X}_{\text{u}}, \mathbf{y}_{\text{u}}$);\\
\While{not converged} {
	\ForEach{censored sample $i$} {
		$\mu,\sigma^2$ := EPM.predict($\mathbf{X}_{\text{c}}^{(i)}$);\\
		$\mathbf{y}_{\text{imp}}^{(i)}$ := mean of $\mathcal{N}(\mu, \sigma^2)_{\geq \mathbf{y}_{\text{c}}^{(i)}}$;\\
	}
	EPM.fit($\mathbf{X}_{\text{u}} || \mathbf{X}_{\text{c}}, \mathbf{y}_{\text{u}} || \mathbf{y}_{\text{imp}}$);\\
}
\Return{$\mathbf{y}_{\text{imp}}$}
\caption{Imputation of censored data}
\label{algo:imputation}
\end{algorithm}
Next, we refit our EPM using the uncensored data and the newly imputed censored data (Line 6).
We then iterate this process until the algorithm converges or until $10$ iterations have been performed. 
%

\subsection{Handling Randomized Algorithms} 
\label{subsub:qrf}

Many algorithms are randomized (e.g., RFs or stochastic gradient descent in machine learning, or stochastic local search solvers in SAT solving).
In order to properly reflect this in our surrogate benchmarks, we should take this randomization into account in our predictions.
Earlier work on surrogate benchmarks~\citep{eggensperger-aaai15} only considered deterministic algorithms and only predicted means; when these methods are applied to randomized algorithms, the result is a deterministic surrogate that can differ qualitatively from the original benchmark.

Instead, we need to predict the entire distribution $P(Y|X)$ and, when asked to output the performance of a single algorithm run, draw a sample from it.
Unfortunately, we do not know in advance the closed form performance distribution (running time distributions have only been studied in some special cases, such as for certain stochastic local search solvers~\citep{hoos-sls04}---if we knew the running time distributions, we could exploit it in the construction of our EPM~\citep{arbelaez-ictai16a}). Instead, to obtain a general solution, we propose to use quantile regression~\citep{koenker-quantilebook05,takeuchi-jmlr06a}. 
Following \cite{meinshausen-jmlr06a}, the $\alpha$-quantile $Q_\alpha(x)$ is defined by
\begin{equation}
Q_\alpha(x) = \inf \{ y: P(Y \leq y | X = x) \geq \alpha\}.
\end{equation}

Since we already know that random forests are well suited as EPMs~\citep{hutter-aij14a,eggensperger-aaai15},
we use a quantile regression forest~(QRF; \citealp{meinshausen-jmlr06a}) for the quantile regression. 
The QRF is very similar to a RF of regression trees: instead of returning the mean over all labels in the selected leafs of the trees,
it returns a given quantile of these labels, $Q_\alpha(x)$.
In our surrogate benchmarks, when asked to predict a randomized algorithm's performance on $x = \langle \conf, \inst \rangle$ with seed $s$, we use $s$ to randomly sample a quantile $\alpha \in [0,1]$ and simply return $Q_\alpha(\conf,\pi)$.
For a deterministic algorithm, we return the median $Q_{0.5}(\conf,\pi)$.

\section{Experiments for Algorithm Configuration}
\label{sec:experiments_ac}

Next, we report experimental results for surrogates based on QRFs in the general AC setting. 
All experiments were performed on Xeon E5-2650 v2 CPUs with 20MB Cache and 64GB RAM running Ubuntu 14.04 LTS.

In the following, we first describe the benchmarks we used to evaluate our approach.
Then, we report results for the predictive quality of EPMs based on QRFs. 
Finally, we show that these EPMs are useful as surrogate benchmarks,
based on an evaluation of the performance of \paramils{}~\citep{hutter-jair09a}, \roar{} and \smac{}~\citep{hutter-lion11a}, and \irace{}~\citep{lopez-ibanez-orp16} on our surrogate benchmarks and the established AC benchmarks from which our surrogates were derived.
For all experiments, we preprocessed the data as described in Section~\ref{sub:surro_setup}, 
imputed right-censored data as described in Algorithm~\ref{algo:imputation}, 
and then trained a QRF as described in Section~\ref{subsub:qrf}, 
with the logarithm of the penalized average running time (PAR10) serving as the response variable for running time optimization benchmarks.\footnote{PAR10 averages all running times, counting each capped run as having taken $10$ times the running time cutoff $\kappa$~\citep{hutter-jair09a}.}

\subsection{Algorithm Configuration Benchmarks from AClib}
\label{ssec:experiments_ac_datasets}

For our experiments, we drew our benchmarks from the algorithm configuration library, AClib~\citep[see \url{www.aclib.net}]{hutter-lion14a}. 
Our first set of benchmarks involves running time minimization; these 
consist of different instance sets taken from each of four widely studied combinatorial problems (mixed-integer programming (MIP), propositional satisfiability (SAT), AI planning, and answer set programming (ASP)) and one or more different solvers for each of these problems (\cplex{}\footnote{\url{http://www-01.ibm.com/software/commerce/optimization/cplex-optimizer/}}, \lingeling{} by~\citealp{lingelingyalsat}, \probsat{} by \citealp{probSAT}, \minisathack{} by \citealp{minisat_hack}, \clasp{} by \citealp{gebser-ai12a} and \lpg{} by \citealp{gerevini-aips02}). 
We used the training-test splits defined in AClib.
Key characteristics of these benchmarks are provided in Table~\ref{tab:ac_benchmarks_overview}, and the underlying AC scenarios are described in detail in Appendix~\ref{app:descr}.

Since AC is a generalization of HPO, we also generated two surrogate HPO benchmarks, which allows us to situate our new results in the context of previous work~\citep{eggensperger-aaai15}.
In these benchmarks, we optimize for misclassification rate (1 $-$ accuracy) on 10-fold cross-validation on training data ($90\%$ of the data)
and then validate the model trained on all of the training data with the final parameter configurations on held-out test data ($10\%$ of the data).
We consider each cross-validation split to be one instance.
We use pseudo instance features in these benchmarks
by simply assigning the $i$-th split to feature value $i$
and the test data with feature value $k+1$ (i.e., $11$). 
Inspired by the automated machine learning tool \textit{auto-sklearn}~\citep{feurer-nips2015a}
and available at \url{https://bitbucket.org/mlindauer/aclib2},
we configured a \textit{SVM}~\citep{cortes-mlj95} on MNIST\footnote{\url{http://www.openml.org/d/554}} and \textit{xgboost}~\citep{chen-kdd16} on covertype\footnote{\url{http://www.openml.org/d/293}}~\citep{collobert-nc02}.
 
We ran \smac{}, \roar{}, and \paramils{} ten times for each running time optimization scenario in order to collect performance data in regions that are likely to be explored by one or more AC procedures.
For the HPO benchmarks, we additionally ran \irace{} ten times.\footnote{Since \irace{} does not implement an adaptive capping mechanism, its authors recommend that it not be used for runtime minimization.}
We report properties of the resulting datasets in Table~\ref{tab:ac_datasets_overview}.
\begin{table}[tbp]
\centering
\scriptsize
\begin{tabular}{lcccccc}
{ } & \multicolumn{1}{c}{\#parameter}  			& \#instance  	& \multirow{2}{*}{total} 	& \#instances 	& conf 			& cutoff\\
{ } & \multicolumn{1}{l}{ ca./int./co.(cond.)} 	& features   	& { } 					& train/test 	& budget	& $\kappa$\\
\toprule
\regions 		& 63/7/4 (4) 					& 148 			& 222 							& 1\,000/1\,000 & 2			& 10000\\
\rcw 			& 63/7/4 (4) 					& 148 			& 222 							& 495/495 		& 5			& 10000\\
\midrule
\rooks          & 38/30/7 (55) 					& 119 			& 194 							& 484/351		& 2			& 300\\
\cf 			& 137/185/0 (0) 				& 119 			& 441 							& 299/302		& 2			& 300\\
\probsatseven	& 5/1/3 (5) 					& 138 			& 128 							& 250/250		& 2			& 300\\
\minisatk 		& 10/0/0 (0) 					& 119 			& 129 							& 300/250		& 2			& 300\\
\midrule
\satellite  	& 48/5/14 (22) 					& 305 			& 372 							& 2\,000/2\,000	& 2			& 300\\
\zeno 			& 48/5/14 (22) 					& 305 			& 372 							& 2\,000/2\,000 & 2			& 300\\
\midrule
\ws 			& 61/30/7 (63) 					& 38 			& 136 							& 240/240		& 4			& 900\\
\midrule
\svmmnist		& 2/1/3 (2)						& 1				& 7								& 10/1			& 500		& 1000\\
\xgboostcovertype & 0/2/9 (0)					& 1				& 12							& 10/1			& 500		& 1000\\
\bottomrule
\end{tabular}
\caption[]{Properties of our AC benchmarks. 
We report the size of the configuration spaces $\pcs$ for the different kinds of parameters (i.e., categorical, integer-valued, continuous and conditionals),
the number of instances features, the total number of input features for our EPMs (\#parameters + \#features), the number of training and test instances, the configuration budget for each AC procedure run (in days for combinatorial problems and number of function evaluations for HPO benchmarks) and the running time cutoff for each target algorithm run (in seconds).
\label{tab:ac_benchmarks_overview}} 
\end{table}
As described in Section~\ref{sub:data}, we used two different setups to collect training data for our EPMs.
Due to memory limitations on our machines, we used at most $1$ million data points to train our EPMs. If we collected more than $1$ million points, we subsampled them to $1$ million.
\begin{table}[tbp]
\centering
\begin{tabular}{l c@{\hskip 1mm}c@{\hskip 1mm}c c@{\hskip 1mm}c@{\hskip 1mm}c@{\hskip 0.5cm}c} 
{} & \multicolumn{3}{c}{\textbf{I}} & \multicolumn{3}{c}{\textbf{II}} & \\ 
{} & $\frac{\#\text{data}}{1000}$ & \%cen & \%to & $\frac{\#\text{data}}{1000}$ & \%cen & \%to  & $\frac{\#\text{conf}}{1000}$ \\ 
\toprule
\regions{}   & 656 & 27 & 0 & 825 & 22 & $<$1 & 198 \\ 
\rcw{}       & 166 & 38 & 2 & 217 & 29 &    1 & 80  \\ 
\midrule
\rooks{}     & 245 & 14 & 3 & 310 & 11 & 5 & 63 \\ 
\cf{}        & 149 & 21 & 9 & 176 & 17 & 9 & 66 \\ 
\probsat{}	 & 199 & 17 & 3 & 245 & 14 & 3 & 28 \\ 
\minisatk{}	 & 177 & 16 & 2 & 218 & 13 & 2 & 37 \\ 
\midrule
\satellite{} & 565 & 27 & $<$1 &   969 & 16 & $<$1 & 252 \\ 
\zeno{}      & 685 & 26 & $<$1 & $>$1K & 17 &    1 & 204 \\ 
\midrule
\ws{}        & 172 & 17 & 4 & 207 & 14 & 4 & 56 \\ 
\midrule
\svmmnist	      & 29 & - & 53 & 29 & - & 52 & 19 \\ 
\xgboostcovertype & 30 & - &  4 & 30 & - &  2 & 22 \\ 
\bottomrule
\end{tabular}
\caption{Properties of our datasets. We list the rounded number$/1000$ of collected $\langle \conf, \inst\rangle$ pairs of 10 runs of each AC procedure, the ratio of right-censored runs (\#cen) and timeouts (\#to) for each of our settings: I, II. We also report the total number$/1000$ of different configurations (\#conf).\label{tab:ac_datasets_overview}
}
\end{table}
%

\subsection{Evaluation of Raw Model Performance}
\label{ssec:experiments_ac_regression}
We now report the predictive performance of QRFs as EPMs.
\cite{hutter-aij14a} performed a similar analysis using random forests as EPMs.
However, their training data differed substantially from ours.
In particular, they uniformly sampled sets of target algorithm configurations and problem instances, and then gathered a performance observation for every entry in the Cartesian product of these sets.
In contrast, we use AC procedures to bias training data towards high-performance regions of the given configuration space; this results in a larger number of configurations in our training data, many of which are evaluated only on few instances.
The question of whether effective EPMs can be trained using such sparse and biased data has not previously been studied
and is an essential requirement for inclusion in our surrogate benchmarks. 

In Table~\ref{tab:exp_ac_scatter_regression},
we show the predictive accuracy of our trained EPMs based on root mean squared error (RMSE; in log space for running time benchmarks) 
to estimate how far our predictions are from true performance values, and Spearman's rank correlation coefficient (CC; \citealp{spearman-ajp04})
to assess whether we can accurately rank different configurations based on predicted performance values.
The latter metric is particularly useful in the context of surrogate benchmarks,
because an AC procedure can make correct decisions as long as the \emph{ranking} of configurations is correct, i.e., the EPM predicts poorly performing configurations to be bad and strong configurations to be good.
To obtain an unbiased estimate of generalization performance,
we used a leave-one-run-out validation splitting scheme:
in each split, we used all but one run of each AC procedure as training data
and evaluated the EPM trained on this data on the remaining runs.
All AC procedures are randomized,
and each AC procedure run is independently initialized with a different random seed.
Therefore, all data points evaluated by a single AC run 
are independent of the points of a different AC run,
even though the two runs may contain identical data points.

Table~\ref{tab:exp_ac_scatter_regression} shows our results on held out data, 
specifically all data collected while configuring the target algorithm on $\insts_{\text{Train}}$ as well as on all data collected during the validation of the incumbent configurations on $\insts_{\text{Test}}$.
As expected, the predictive performance of our EPM on $\insts_{\text{Train}}$ is quite similar between Setting I and II.
However on $\insts_{\text{Test}}$, 
Setting II performed significantly better than Setting I across our benchmarks (p-values of $0.0021$ on RMSE and $0.0067$ on CC based on a one-sided, non-parametric permutation test, cf.~\citealp{hoos-emp17a}).
\begin{table}[tbp]
\center
\begin{tabular}{@{\hskip 1mm}l cc cc c cc cc@{\hskip 1mm}}
{} & \multicolumn{4}{c}{\textbf{RMSE}} && \multicolumn{4}{c}{\textbf{CC}} \\
{} & \multicolumn{2}{l}{\textbf{Configuration}} & \multicolumn{2}{l}{\textbf{Validation}} & & \multicolumn{2}{l}{\textbf{Configuration}} & \multicolumn{2}{l}{\textbf{Validation}}  \\
              & \textbf{I} & \textbf{II} & \textbf{I} & \textbf{II} & & \textbf{I} & \textbf{II} & \textbf{I} & \textbf{II} \\
 \toprule
 \regions{}          &  0.2~  &  0.19   &  0.33  &  0.2~  & & 0.92 & 0.92  & 0.67 & 0.9~ \\
 \rcw{}              &  0.12  &  0.12   &  0.08  &  0.08  & & 0.98 & 0.98  & 0.99 & 0.99 \\
 \midrule 
 \rooks              &  0.35  &  0.35   &  0.49  &  0.42  & & 0.98 & 0.98  & 0.98 & 0.98 \\
 \cf{}               &  0.35  &  0.35   &  0.72  &  0.31  & & 0.86 & 0.86  & 0.7~ & 0.93 \\
 \probsatseven{}     &  0.6~  &  0.6~   &  0.82  &  0.54  & & 0.69 & 0.69  & 0.36 & 0.78 \\
 \minisatk{}         &  0.27  &  0.27   &  0.48  &  0.24  & & 0.96 & 0.96  & 0.89 & 0.97 \\
 \midrule 
 \satellite{}        &  0.1~  &  0.1~   &  0.14  &  0.13  & & 0.8~ & 0.8~  & 0.92 & 0.92 \\
 \zeno{}             &  0.27  &  0.29   &  0.38  &  0.39  & & 0.7~ & 0.69  & 0.78 & 0.77 \\
 \midrule 
 \ws{}               &  0.31  &  0.31   &  0.64  &  0.42  & & 0.95 & 0.95  & 0.85 & 0.95 \\
 \midrule 
 \svmmnist{}         &  0.06  &  0.06   &  0.06  &  0.02  & & 0.99 & 0.99  & 0.7~ & 0.75 \\
 \xgboostcovertype{} &  0.25  &  0.26   &  0.17  &  0.14  & & 0.87 & 0.85  & 0.85 & 0.85 \\
 \bottomrule
\end{tabular}
\caption{Leave-one-run-out model performance. We report mean root mean squared error (RMSE) and Spearman's rank correlation coefficient (CC) of log PAR10 running times (for AC scenarios) and loss (for HPO scenarios) across ten runs for which the EPM was trained using data from setting I or II (see Section~\ref{ssec:experiments_ac_datasets}). For each run, we trained an EPM on all but one configuration run for each considered AC procedure and report average results across left-out runs. Using the QRF, we predicted the median. We report results on all data collected during configuring the target algorithm on $\inst_{\text{Train}}$ and the data collected during the validation of the incumbent configurations on $\inst_{\text{Test}}$.\label{tab:exp_ac_scatter_regression}}
\end{table}
Overall, our EPMs yielded rather accurate target algorithm performance predictions, and achieved high overall correlation ($CC \ge 0.75$) in $9$ out of $11$ benchmarks wrt $\insts_{\text{Train}}$ and in all benchmarks wrt $\insts_{\text{Test}}$ using Setting II.
The RMSE on the running time benchmarks was substantially smaller than $1.0$, i.e., the predictions are less than one order of magnitude off on average.
Considering the HPO benchmarks, our models were more accurate for \svmmnist{} than they were for \xgboostcovertype{}. This difference was driven by timeouts (counted using the maximal error value of $1$).
For \svmmnist{}, these timeouts were easier to predict (mostly driven by a small number of parameters); in contrast, a potential timeout can depend on more complex interactions of parameters in the case of \xgboostcovertype{}. 
The predictions for non-timeout runs for \xgboostcovertype{} were roughly as good as for \svmmnist. Since Setting II performed consistently better than Setting I,
in the following we consider only Setting II.
%
\footnote{
To study whether it is necessary for our model to distinguish different
instances, we also considered another baseline, inspired by a metric
used by~\citealp{soares-mlj04a}:
We computed the rank correlation between the true running times and the mean
running times per configuration across instances (aka mean regressor).
A low correlation coefficient indicates that the instances differ in hardness
or that the rank of configurations changes between instances. Indeed, for most
of our benchmarks we obtained a low correlation coefficient $CC \le 0.35$,
indicating that it is necessary for the model to consider instances to obtain
accurate predictions. For \svmmnist{}, \xgboostcovertype{}, \satellite{}, and
\zeno{}, we obtained $CC \ge 0.79$ for data points used during validation,
showing that the instances in these benchmarks are rather similar.
(We note that these numbers are based on observed runs and not for predicting
performance on unseen instances or configurations.)
}

\subsection{Evaluation of Surrogates as Benchmarks for Algorithm Configuration}
\label{ssec:exp_ac_surrogates}

We now turn to the most important experimental question: how well our QRF-based EPMs work as surrogate benchmarks for algorithm configuration procedures. 
For these experiments, we trained and saved a QRF model on the imputed data from our Setting II (described above). To evaluate these EPMs as AC benchmarks, we re-ran our configuration experiments, now obtaining running time measurements from an EPM (running as a background process) rather than the real target algorithm. 
Doing so reduced the average 
CPU time required for evaluating a configuration on a single instance from $27.29 \pm 100.26$ ($\mu \pm \sigma$) seconds to $0.23 \pm 0.13$ seconds. 

We also considered \textit{leave-one-configurator-out (\loco{})} evaluations, training each EPM on data gathered by all but one AC procedure, and then running the remaining AC procedures on this surrogate benchmark to simulate benchmarking a new AC procedure.
	
\begin{figure}[tbp]
	\begin{tabular}{c c c}
		\small{\textbf{Original} Benchmark} & \small{\textbf{QRF} Surrogate Benchmark} & \small{\textbf{QRF} Surrogate Benchmark} \\
		{ } & \small{All Data/Setting II} & \small{\loco{}/Setting II} \\
		\toprule
		\multicolumn{3}{l}{\textbf{\rcw{}}} \\
		\raisebox{-0.7\totalheight}{\includegraphics[width=0.28\textwidth]{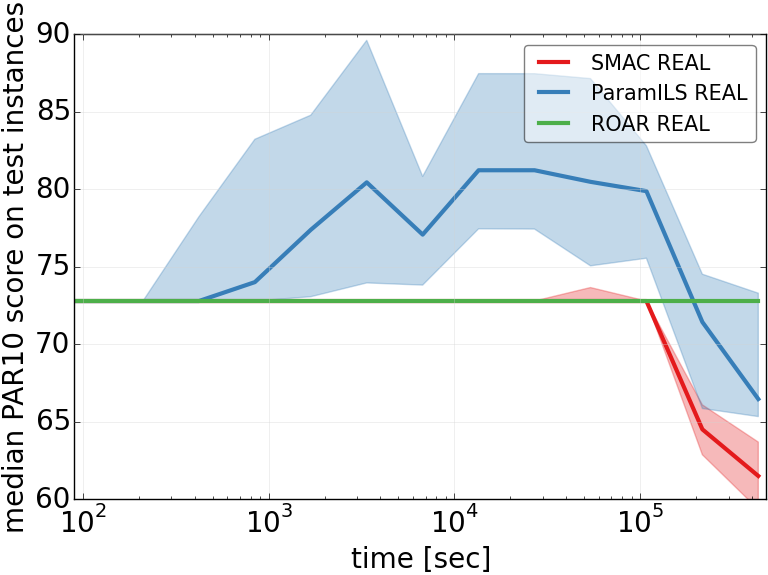}}
		& \raisebox{-0.7\totalheight}{\includegraphics[width=0.28\textwidth]{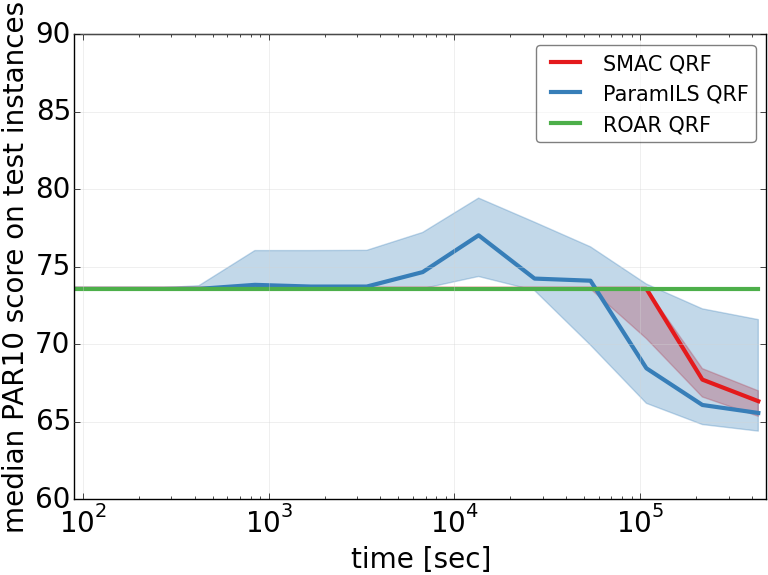}}
		& \raisebox{-0.7\totalheight}{\includegraphics[width=0.28\textwidth]{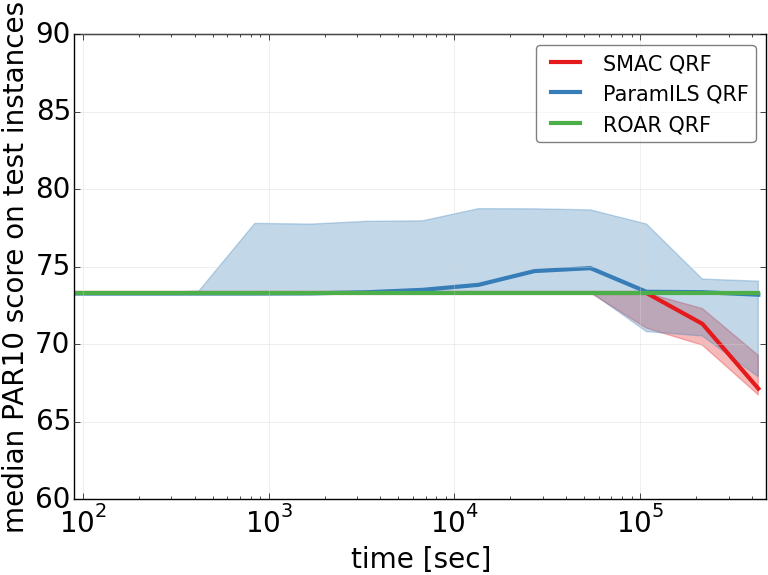}}
		\\
		\midrule
		\multicolumn{3}{l}{\textbf{\rooks{}}} \\
		\raisebox{-0.7\totalheight}{\includegraphics[width=0.28\textwidth]{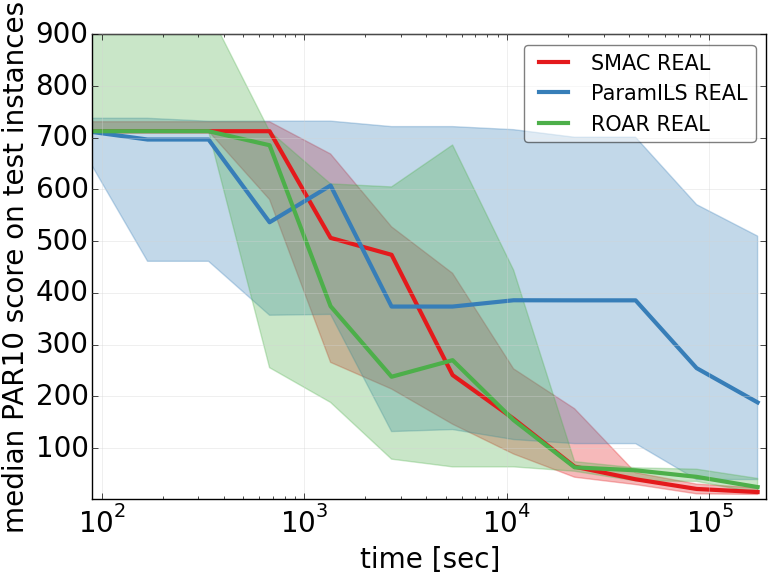}}
		& \raisebox{-0.7\totalheight}{\includegraphics[width=0.28\textwidth]{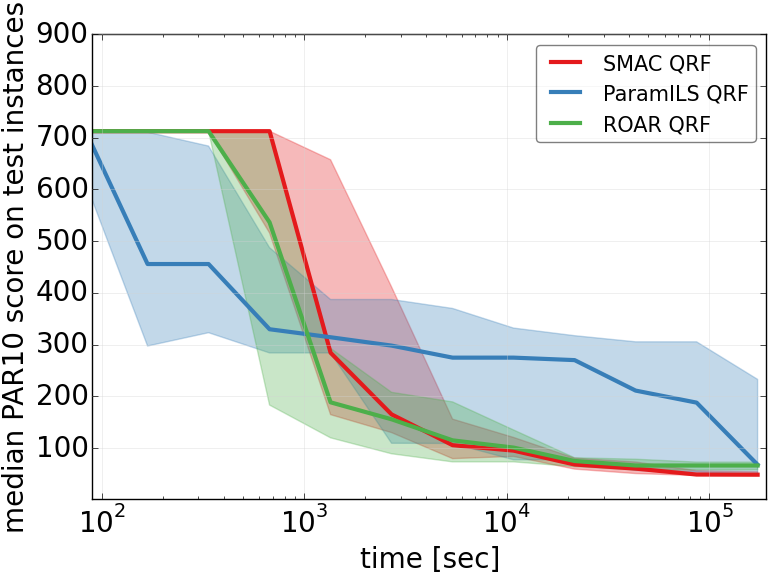}}
		& \raisebox{-0.7\totalheight}{\includegraphics[width=0.28\textwidth]{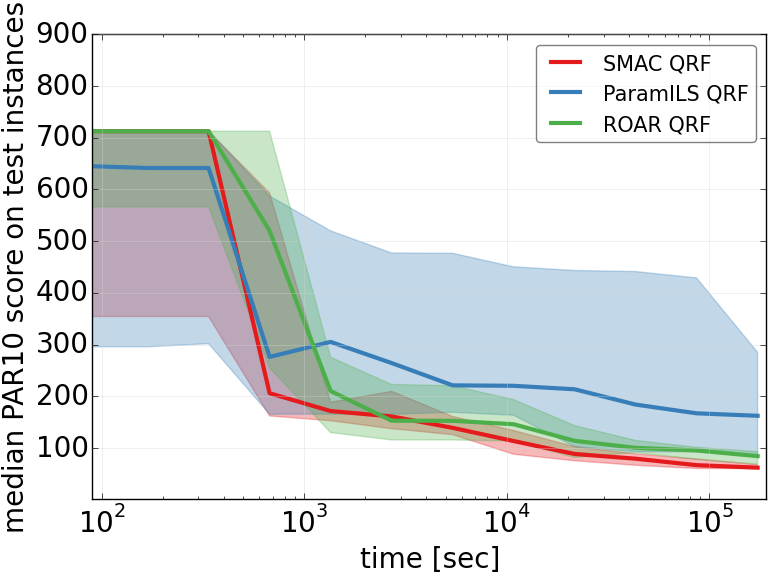}}
		\\
		\midrule
		\multicolumn{3}{l}{\textbf{\zeno{}}} \\
		\raisebox{-0.7\totalheight}{\includegraphics[width=0.28\textwidth]{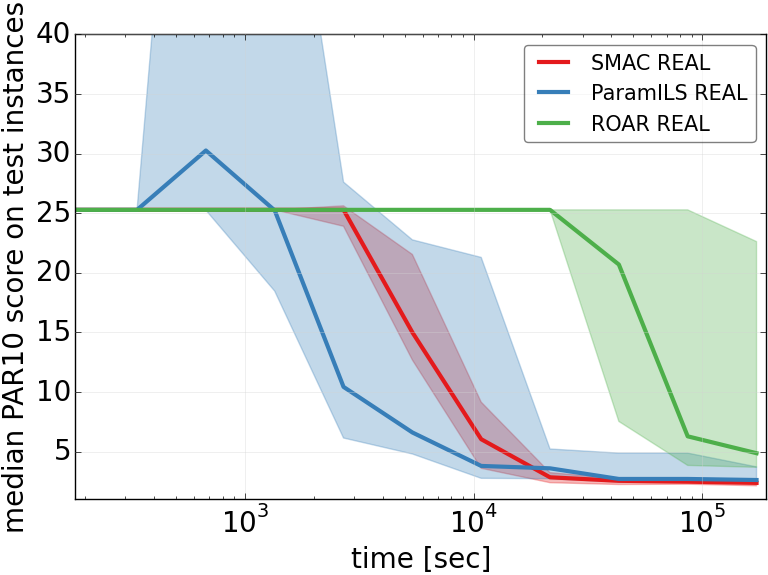}}
		& \raisebox{-0.7\totalheight}{\includegraphics[width=0.28\textwidth]{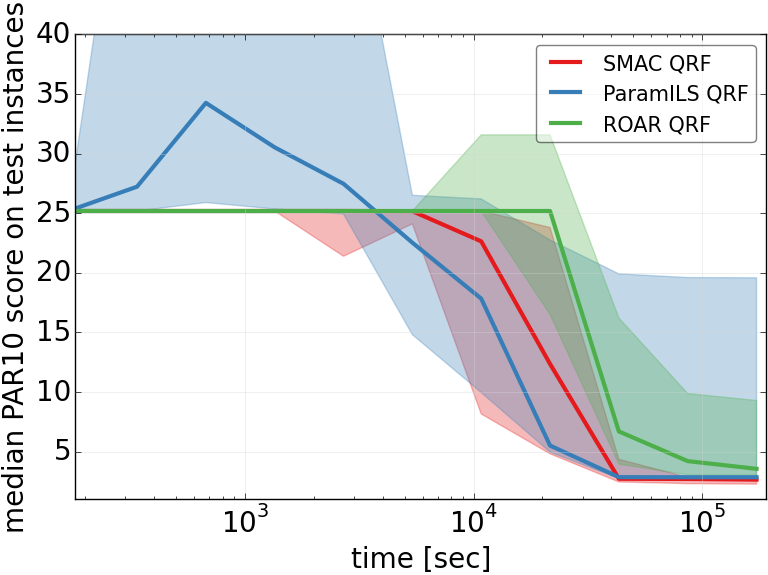}}
		& \raisebox{-0.7\totalheight}{\includegraphics[width=0.28\textwidth]{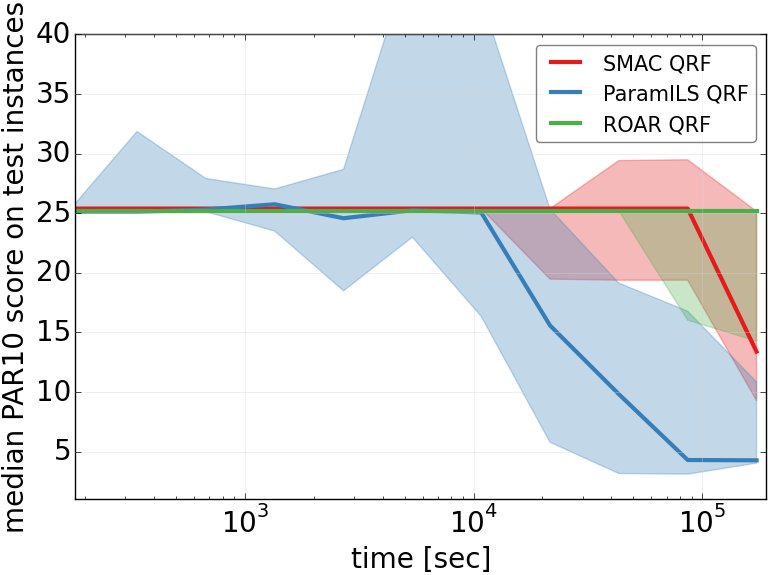}} \\
		\\
		\midrule
		\multicolumn{3}{l}{\textbf{\ws{}}} \\
		\raisebox{-0.7\totalheight}{\includegraphics[width=0.28\textwidth]{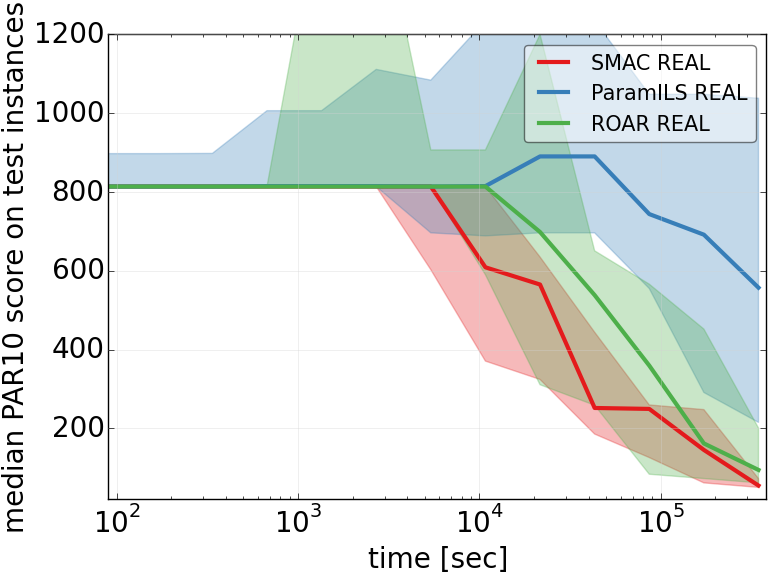}}
		& \raisebox{-0.7\totalheight}{\includegraphics[width=0.28\textwidth]{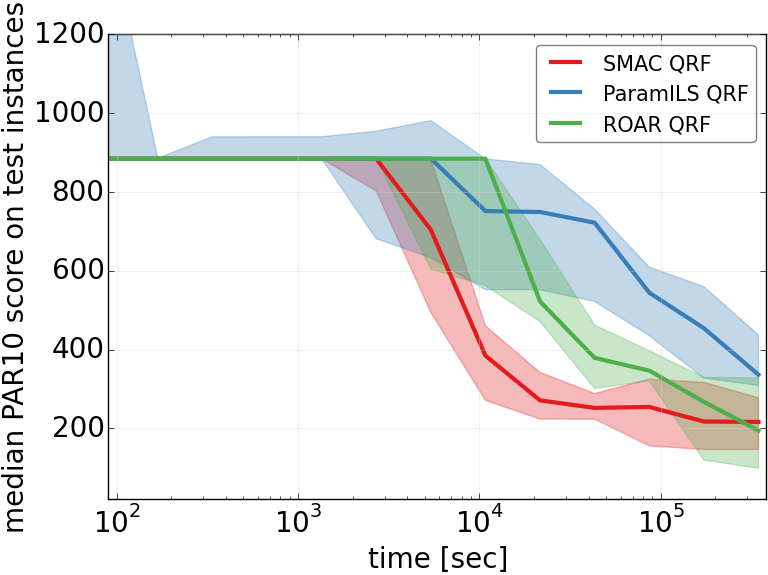}}
		& \raisebox{-0.7\totalheight}{\includegraphics[width=0.28\textwidth]{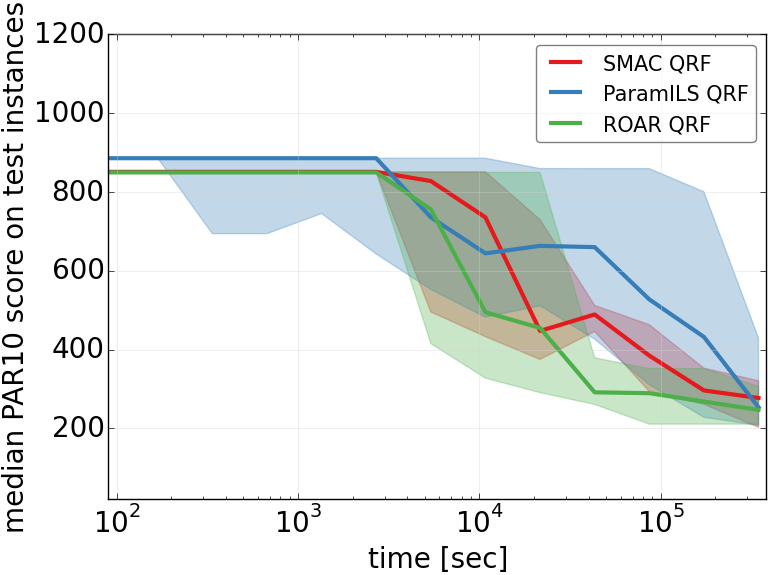}} \\
        \\
        \midrule
		\multicolumn{3}{l}{\textbf{\svmmnist{}}} \\
		\raisebox{-0.7\totalheight}{\includegraphics[width=0.28\textwidth]{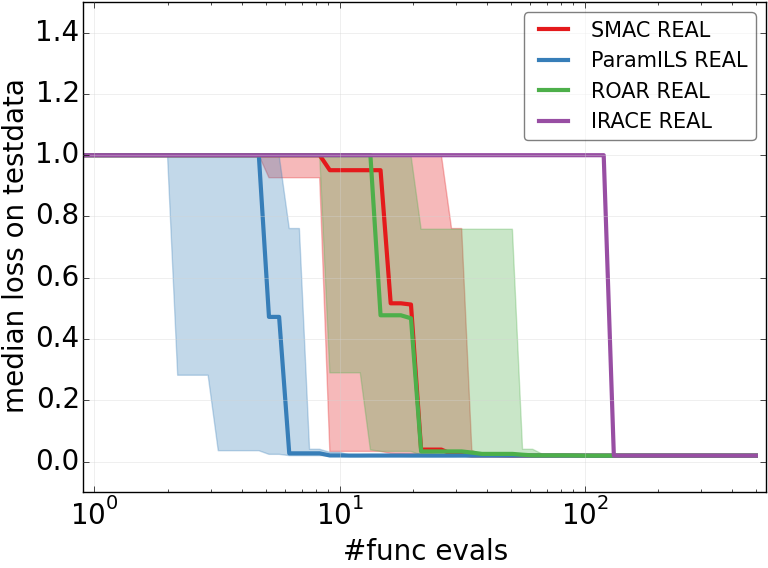}}
		& \raisebox{-0.7\totalheight}{\includegraphics[width=0.28\textwidth]{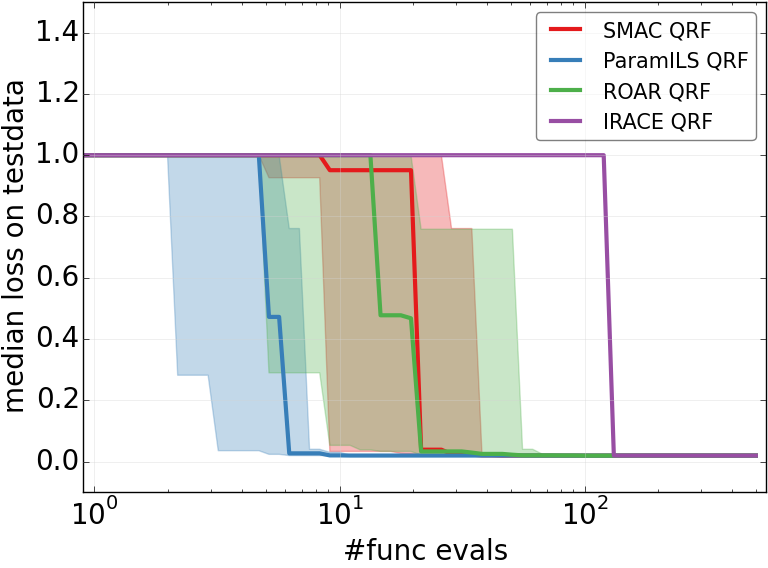}}
		& \raisebox{-0.7\totalheight}{\includegraphics[width=0.28\textwidth]{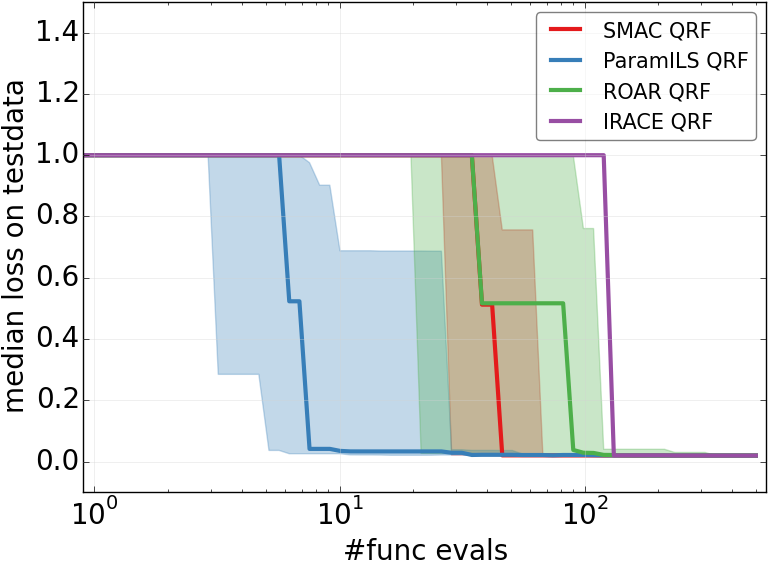}} \\
	\end{tabular}
	\caption{Best performance found by different AC procedures over time. We plot median and quartile of best performance across 10 runs of each AC procedure over time (for \svmmnist{} we use number of function evaluations) on the original benchmark (left) and on QRF-based surrogate benchmarks trained either on data from \textbf{all AC procedures} (middle) or \textbf{leave-one-configurator-out} data (right). \label{fig:exp_ac_trajectories}}
\end{figure}

When used in the context of AC benchmarks, the absolute quality of running time
predictions is less important than the ranking of the AC procedures. Therefore we study performance as a function of time in Figure~\ref{fig:exp_ac_trajectories} to visually compare the behaviour of different AC procedures on the original and surrogate-base benchmarks. We observe that the relative rankings between AC procedures were well preserved for surrogates trained on all data: \smac{} was correctly predicted to outperform \roar{} in all AC scenarios and at almost all time steps.
\paramils{}'s performance was predicted slightly less well, but ranks were still preserved well across scenarios and time steps.
Also, the surrogate-based benchmarks captured overtuning effects as present in \zeno{} and \rcw{}. For the machine learning benchmarks, we obtained almost perfect surrogate benchmarks, with \irace{} and \paramils{} performing similarly, although \paramils{} having a slightly higher inter-quartile ratio than on the original benchmark.

In the \loco{} setting (Figure~\ref{fig:exp_ac_trajectories}, right
column), the relative performance of \smac{} and \roar{} was still predicted
correctly throughout except for the \zeno{} benchmark, where \smac{} and \roar{}
did not improve as much as on the original benchmark. \paramils{}, again, was
predicted slightly worse, but its relative ranks were still predicted
correctly, with two exceptions: On the \loco{} surrogate of benchmark \rcw{},
\paramils{} performed worse than on the original benchmark (and could not find a
configuration better than the default), and on the \loco{} surrogate of
benchmark \ws{}, \paramils{} performed better than on the original
benchmark.
We believe that this is due to the substantial differences in search strategies between \paramils{} and the other AC procedures, leading to qualitatively different sets of performance data and hence EPMs; surrogate benchmarks constructed based on data from global search algorithms intuitively capture areas of weak/strong performance well, but do not necessarily capture fine local variation and may thus (as discussed above) not work as well for gradient-following AC procedures.

To also provide a quantitative evaluation of how closely our surrogate
benchmarks resemble the original benchmarks, we used an error metric based on
the idea that a surrogate benchmark should preserve the outcomes of pairwise
comparisons of AC procedures obtained on the underlying original benchmarks,
across different overall running time budgets.
To deal with performance variability due to randomization in target algorithm
and AC procedure runs, we applied statistical tests to determine whether one AC
procedure performed significantly better than another.
For each running time budget (number of target algorithm or surrogate
evaluations, resp.) starting from $\kappa$ (or $2$), we used a
Kruskal-Wallis-Test and a pairwise post-hoc Wilcoxon rank-sum test with
Bonferroni's multiple testing correction ($\alpha=0.05$).
Table~\ref{tab:error_metric} shows how our metric penalizes differences
in the outcomes of this statistical test between the original and the
surrogate versions of a benchmark. This metric was inspired by
\cite{leite-ecai10a}, but we additionally penalized the case in which true
performance values do not differ significantly while our surrogate-based
predictions do. To obtain our overall error values, we averaged across the
values of this metric for each pair of AC procedures in our comparison and then
averaged the error values thus obtained over the different time budgets
considered. We note that this metric provides a quantitative measure of the
similarity of the qualitative trajectory diagrams shown in
Figure~\ref{fig:exp_ac_trajectories}.
\begin{table}[tbp]
\centering
\begin{tabular}{c | ccc}
original $\backslash$ surrogate	& better & equal & worse\\
\toprule
better & $0$ & $0.5$ & $1$ \\
equal & $0.5$ & $0$  & $0.5$\\
worse & $1$	& $0.5$	 & $0$\\
\bottomrule
\end{tabular}
\caption{Overview of our error metric quantifying the degree to which using surrogates in performance comparisons between two given configurators yields results statistically significant from those obtained based on the underlying original benchmarks.
\label{tab:error_metric}}
\end{table}

In Table~\ref{tab:ac_error}, we report this metric for both surrogates
based on \textit{all data} and for the \loco{} setting.
All our surrogate benchmarks achieved an error lower than $0.5$, which indicates that, on average, using our surrogates produces behaviour qualitatively similar to that observed for the underlying target algorithms.
In most \loco{} experiments, we observed slightly higher error values (but still below 0.5), because our EPMs have never seen data from the AC procedure that was run on the respective surrogate benchmark.
On the \probsat{} scenario, the AC procedures running on surrogates show
qualitatively similar behaviour as on the original benchmark and achieved an
error of $0$, although the EPMs were relatively weak (comparatively high RMSE
and low CC; see Table~\ref{tab:exp_ac_scatter_regression}). This occurred
because \probsat{} only has a few important parameters, which all AC procedures
identified on the original and surrogate benchmarks. For \regions{},
Table~\ref{tab:ac_error} reports the highest difference between the error on
all data and the \loco{} setting. In our experiments we observed that for this
benchmark, \smac{} performed merely en par with \roar{} in the \loco{}
setting, whereas it performed substantially better on the original benchmark.
This resulted in a high value of our metric.
A similar phenomenon was observed on \xgboostcovertype{}, where \paramils{}
found a significantly better-performing configuration earlier and therefore
constantly added to the error. Overall, our results indicate that the relative
performance of AC procedures on surrogate-based benchmarks largely resembles
that observed on benchmarks involving much costlier target algorithm runs, but
that the resemblance is higher when our surrogates are trained based
on all available data.
\begin{table}[tbp]
    \begin{center}
    \begin{minipage}[t]{0.48\columnwidth} 
	\begin{tabular}{lll}
		\multirow{2}{*}{\textbf{Scenario}} & \multirow{2}{*}{\textbf{All Data}} & \multirow{2}{*}{\textbf{\loco{}}} \\
		{} & {} & {} \\
 \toprule
  \regions{}      &  0.17  &   0.47 \\
  \rcw{}          &  0.08 &   0.25 \\
 \midrule
  \rooks{}        &  0.07 &   0.05 \\
  \cf{}           &  0.1   &   0.12 \\
  \probsatseven{} &  0     &   0.17 \\
  \minisatk{}     &  0.02 &   0    \\
 \midrule
  \satellite{}    &  0     &   0.05 \\
  \zeno{}         &  0.1   &   0.12 \\
 \midrule
  \ws{}           &  0.07 &   0.15 \\
 \midrule
  \svmmnist{}         & 0.01 &  0.05 \\
  \xgboostcovertype{} & 0.18   &  0.21  \\
 \bottomrule
	\end{tabular}
	\end{minipage}
    \begin{minipage}[t]{0.35\columnwidth}
    \centering
	\begin{tabular}{ll}
    \textbf{Mean}         & \textbf{Speed} \\
	\textbf{Running Time} &  \textbf{Up}   \\
	\toprule
    ~7.3  & ~36 \\
    ~82.1 & 497   \\
    \midrule
    ~21.3 & 128   \\
    ~36   & 199   \\
    ~26.5 & 159   \\
    ~30.1 & 183   \\
    \midrule
    ~8.21 & ~34 \\
    ~6.79 & ~22 \\
    \midrule
    ~61.4 & 383   \\
    \midrule
    658   & 1641 \\ 
    566   & 1434 \\ 
	\bottomrule
	\end{tabular}
    \end{minipage}   
    \end{center}    
\caption{Error of our surrogate-based benchmarks and speedups provided. We report the weighted average difference between pairs of AC procedures averaged over time (\textbf{left}) and the mean running time (in CPU seconds) across all evaluated real target algorithm runs with the speedup gained when using surrogate benchmarks (\textbf{right}). One prediction took on average $0.23 \pm 0.13$ seconds.\label{tab:ac_error}}
\end{table}

In Table~\ref{tab:ac_error}, we also report average speedups gained per target
algorithm run. Our surrogate benchmarks allow dramatic speedups in
experimentation, cutting down the time required for algorithm configuration by
a factor of over $1000$ for the most expensive AC benchmarks.
This will substantially ease the development of AC procedures by facilitating unit testing, debugging, and whitebox testing. 
Furthermore, the behaviour of AC procedures on standard benchmarks involving actual target algorithm runs is captured closely enough by our surrogate-based benchmarks that it makes sense to use the latter in comparative performance evaluations 
of AC procedures.

\section{Conclusion}
\label{sec:conclusion}

We presented a novel approach for constructing model-based surrogate benchmarks for the general problem of AC---subsuming HPO.
Our surrogate benchmarks replace expensive evaluations of algorithm configurations by cheap performance predictions based on EPMs with speedups in excess of up to four orders of magnitude.
Our efficient surrogate benchmarks can (i) substantially speed up debugging 
and unit testing of AC procedures,
(ii) facilitate white-box testing,
and (iii) provide a basis for assessing and comparing AC procedure performance.

To construct EPMs for using them as surrogates in AC benchmarks,
we proposed to use AC procedures to generate training data for our EPM
to focus on the more relevant high-performance regions of the parameter configuration space.
We further addressed challenges of AC by studying different ways to collect data on training and test instances,
imputation of right-censored data
and predicting performance of randomized algorithms; latter by introducing EPMs based on quantile regression forests.

In comprehensive experiments with benchmarks from AClib,
we showed that our surrogate benchmarks are well able to stand in for AC benchmarks.
An issue arises from large amounts of target algorithm performance data; for some of our AC benchmarks, 
we had over $1$ million data points available, which we subsequently had to subsample to avoid memory issues in the construction of EPMs; however, better solutions to this problem can likely be devised.
Since deep neural networks have recently shown impressive results for big data sets and natively support training in batches, we 
plan to study scalable Bayesian neural networks~\citep{neal-phd95,blundell-icml15a,snoek-icml15a,springenberg-nips16a}
to predict the performance of randomized algorithms.

\begin{acknowledgements}

We thank Stefan Falkner for the implementation of the quantile regression forest used in our experiments and for fruitful discussions on early drafts of the paper.  
K.\ Eggensperger, M.\ Lindauer and F.\ Hutter acknowledge funding by the DFG (German Research Foundation) under Emmy Noether grant HU 1900/2-1; 
K.\ Eggensperger also acknowledges funding by the State Graduate Funding Program of Baden-Württemberg.
H.\ Hoos and K.\ Leyton-Brown acknowledge funding through NSERC Discovery Grants; K.\ Leyton-Brown also acknowledges funding from an NSERC E.W.R.\ Steacie Fellowship.

\end{acknowledgements}

\bibliographystyle{spbasic} 
\bibliography{strings,lib,local,proc}

\begin{thebibliography}{104}
\providecommand{\natexlab}[1]{#1}
\providecommand{\url}[1]{{#1}}
\providecommand{\urlprefix}{URL }
\expandafter\ifx\csname urlstyle\endcsname\relax
  \providecommand{\doi}[1]{DOI~\discretionary{}{}{}#1}\else
  \providecommand{\doi}{DOI~\discretionary{}{}{}\begingroup
  \urlstyle{rm}\Url}\fi
\providecommand{\eprint}[2][]{\url{#2}}

\bibitem[{Ahmadizadeh et~al(2010)Ahmadizadeh, Dilkina, Gomes, and
  Sabharwal}]{ahmadizadeh-cp10a}
Ahmadizadeh K, Dilkina B, Gomes C, Sabharwal A (2010) An empirical study of
  optimization for maximizing diffusion in networks. In: Cohen D (ed)
  Proceedings of the International Conference on Principles and Practice of
  Constraint Programming (CP'10), Springer-Verlag, Lecture Notes in Computer
  Science, vol 6308, pp 514--521

\bibitem[{Ansel et~al(2014)Ansel, Kamil, Veeramachaneni, Ragan{-}Kelley,
  Bosboom, O'Reilly, and Amarasinghe}]{ansel-pact14a}
Ansel J, Kamil S, Veeramachaneni K, Ragan{-}Kelley J, Bosboom J, O'Reilly U,
  Amarasinghe S (2014) Opentuner: an extensible framework for program
  autotuning. In: Amaral J, Torrellas J (eds) International Conference on
  Parallel Architectures and Compilation, {ACM}, pp 303--316

\bibitem[{Ans{\'o}tegui et~al(2009)Ans{\'o}tegui, Sellmann, and
  Tierney}]{ansotegui-cp09a}
Ans{\'o}tegui C, Sellmann M, Tierney K (2009) A gender-based genetic algorithm
  for the automatic configuration of algorithms. In: Gent I (ed) Proceedings of
  the Fifteenth International Conference on Principles and Practice of
  Constraint Programming (CP'09), Springer-Verlag, Lecture Notes in Computer
  Science, vol 5732, pp 142--157

\bibitem[{Ans{\'o}tegui et~al(2015)Ans{\'o}tegui, Malitsky, Sellmann, and
  Tierney}]{ansotegui-ijcai15a}
Ans{\'o}tegui C, Malitsky Y, Sellmann M, Tierney K (2015) Model-based genetic
  algorithms for algorithm configuration. In:  \cite{ijcai15}, pp 733--739

\bibitem[{Arbelaez et~al(2016)Arbelaez, Truchet, and
  O'Sullivan}]{arbelaez-ictai16a}
Arbelaez A, Truchet C, O'Sullivan B (2016) Learning sequential and parallel
  runtime distributions for randomized algorithms. In: Proceedings of the
  international conference on tools with artificial intelligence (ICTAI)

\bibitem[{Bach and Blei(2015)}]{icml15}
Bach F, Blei D (eds) (2015) Proceedings of the 32nd International Conference on
  Machine Learning (ICML'15), vol~37, Omnipress

\bibitem[{Balint and Sch{\"o}ning(2012)}]{probSAT}
Balint A, Sch{\"o}ning U (2012) Choosing probability distributions for
  stochastic local search and the role of make versus break. In:  \cite{sat12},
  pp 16--29

\bibitem[{Bardenet et~al(2014)Bardenet, Brendel, K{\'e}gl, and
  Sebag}]{bardenet-icml13a}
Bardenet R, Brendel M, K{\'e}gl B, Sebag M (2014) Collaborative hyperparameter
  tuning. In:  \cite{icml13}, pp 199--207

\bibitem[{Bartlett et~al(2012)Bartlett, Pereira, Burges, Bottou, and
  Weinberger}]{nips12}
Bartlett P, Pereira F, Burges C, Bottou L, Weinberger K (eds) (2012)
  Proceedings of the 26th International Conference on Advances in Neural
  Information Processing Systems (NIPS'12)

\bibitem[{Belov et~al(2014)Belov, Diepold, Heule, and
  J{\"a}rvisalo}]{satchallenge14}
Belov A, Diepold D, Heule M, J{\"a}rvisalo M (eds) (2014) Proceedings of {SAT}
  Competition 2014: Solver and Benchmark Descriptions, Department of Computer
  Science Series of Publications B, vol B-2014-2, University of Helsinki

\bibitem[{Bensusan and Kalousis(2001)}]{bensusan_ecml01}
Bensusan H, Kalousis A (2001) Estimating the predictive accuracy of a
  classifier. In: Proceedings of the 12th European Conference on Machine
  Learning (ECML), Springer, pp 25--36

\bibitem[{Bergstra et~al(2014)Bergstra, Yamins, and Cox}]{bergstra-icml13a}
Bergstra J, Yamins D, Cox D (2014) Making a science of model search:
  Hyperparameter optimization in hundreds of dimensions for vision
  architectures. In:  \cite{icml13}, pp 115--123

\bibitem[{Biere(2013)}]{biere-tech13a}
Biere A (2013) Lingeling, plingeling and treengeling entering the sat
  competition 2013. In: Balint A, Belov A, Heule M, J{\"a}rvisalo M (eds)
  Proceedings of {SAT} Competition 2013: Solver and Benchmark Descriptions,
  University of Helsinki, Department of Computer Science Series of Publications
  B, vol B-2013-1, pp 51--52

\bibitem[{Biere(2014)}]{lingelingyalsat}
Biere A (2014) Yet another local search solver and {Lingeling} and friends
  entering the {SAT} competition 2014. In:  \cite{satchallenge14}, pp 39--40

\bibitem[{Birattari et~al(2002)Birattari, Stützle, Paquete, and
  Varrentrapp}]{birattari-gecco02a}
Birattari M, Stützle T, Paquete L, Varrentrapp K (2002) A racing algorithm for
  configuring metaheuristics. In: Langdon W, Cantu-Paz E, Mathias K, Roy R,
  Davis D, Poli R, Balakrishnan K, Honavar V, Rudolph G, Wegener J, Bull L,
  Potter M, Schultz A, Miller J, Burke E, Jonoska N (eds) Proceedings of the
  Genetic and Evolutionary Computation Conference (GECCO'02), Morgan Kaufmann
  Publishers, pp 11--18

\bibitem[{Bischl et~al(2016)Bischl, Kerschke, Kotthoff, Lindauer, Malitsky,
  Frech\'{e}tte, Hoos, Hutter, Leyton-Brown, Tierney, and
  Vanschoren}]{bischl-aij16a}
Bischl B, Kerschke P, Kotthoff L, Lindauer M, Malitsky Y, Frech\'{e}tte A, Hoos
  H, Hutter F, Leyton-Brown K, Tierney K, Vanschoren J (2016) {ASlib}: A
  benchmark library for algorithm selection. Artificial Intelligence pp 41--58

\bibitem[{Blundell et~al(2015)Blundell, Cornebise, Kavukcuoglu, and
  Wierstra}]{blundell-icml15a}
Blundell C, Cornebise J, Kavukcuoglu K, Wierstra D (2015) Weight uncertainty in
  neural network. In:  \cite{icml15}, pp 1613--1622

\bibitem[{Bonet and Koenig(2015)}]{aaai15}
Bonet B, Koenig S (eds) (2015) Proceedings of the Twenty-nineth National
  Conference on Artificial Intelligence (AAAI'15), AAAI Press

\bibitem[{Brazdil et~al(2008)Brazdil, Giraud-Carrier, Soares, and
  Vilalta}]{brazdil-08}
Brazdil P, Giraud-Carrier C, Soares C, Vilalta R (2008) Metalearning:
  Applications to Data Mining, 1st edn. Springer Publishing Company,
  Incorporated

\bibitem[{Breimann(2001)}]{breimann-mlj01a}
Breimann L (2001) Random forests. Machine Learning Journal 45:5--32

\bibitem[{Brochu et~al(2010)Brochu, Cora, and de~Freitas}]{brochu-corr10a}
Brochu E, Cora V, de~Freitas N (2010) A tutorial on {Bayesian} optimization of
  expensive cost functions, with application to active user modeling and
  hierarchical reinforcement learning. Computing Research Repository (CoRR)
  abs/1012.2599

\bibitem[{Brummayer et~al(2012)Brummayer, Lonsing, and
  Biere}]{brummayer-sat10a}
Brummayer R, Lonsing F, Biere A (2012) Automated testing and debugging of {SAT}
  and {QBF} solvers. In:  \cite{sat12}, pp 44--57

\bibitem[{Chen and Guestrin(2016)}]{chen-kdd16}
Chen T, Guestrin C (2016) Xgboost: {A} scalable tree boosting system. In:
  Krishnapuram B, Shah M, Smola A, Aggarwal C, Shen D, Rastogi R (eds)
  Proceedings of the 22nd {ACM} {SIGKDD} International Conference on Knowledge
  Discovery and Data Mining (KDD), {ACM}, pp 785--794

\bibitem[{Cimatti and Sebastiani(2012)}]{sat12}
Cimatti A, Sebastiani R (eds) (2012) Proceedings of the Fifteenth International
  Conference on Theory and Applications of Satisfiability Testing (SAT'12),
  Lecture Notes in Computer Science, vol 7317, Springer-Verlag

\bibitem[{Coelho et~al(2010)Coelho, Studer, and Wooldridge}]{ecai10}
Coelho H, Studer R, Wooldridge M (eds) (2010) Proceedings of the Nineteenth
  European Conference on Artificial Intelligence (ECAI'10), IOS Press

\bibitem[{Collobert et~al(2002)Collobert, Bengio, and Bengio}]{collobert-nc02}
Collobert R, Bengio S, Bengio Y (2002) A parallel mixture of svms for very
  large scale problems. Neural Computation 14(5):1105--1114

\bibitem[{Cortes and Vapnik(1995)}]{cortes-mlj95}
Cortes C, Vapnik V (1995) Support-vector networks. Machine Learning
  20(3):273--297

\bibitem[{Dasgupta and McAllester(2014)}]{icml13}
Dasgupta S, McAllester D (eds) (2014) Proceedings of the 30th International
  Conference on Machine Learning (ICML'13), Omnipress

\bibitem[{Dixon and Szeg{\"o}(1978)}]{dixon1978global}
Dixon L, Szeg{\"o} G (1978) The global optimization problem: an introduction.
  Towards global optimization 2:1--15

\bibitem[{Domhan et~al(2015)Domhan, Springenberg, and Hutter}]{domhan-ijcai15}
Domhan T, Springenberg JT, Hutter F (2015) Speeding up automatic hyperparameter
  optimization of deep neural networks by extrapolation of learning curves. In:
   \cite{ijcai15}, pp 3460--3468

\bibitem[{E{\'{e}}n and S{\"{o}}rensson(2003)}]{een-sat03}
E{\'{e}}n N, S{\"{o}}rensson N (2003) An extensible sat-solver. In: Giunchiglia
  E, Tacchella A (eds) Proceedings of the conference on Theory and Applications
  of Satisfiability Testing (SAT), Springer, Lecture Notes in Computer Science,
  vol 2919, pp 502--518

\bibitem[{Eggensperger et~al(2013)Eggensperger, Feurer, Hutter, Bergstra,
  Snoek, Hoos, and Leyton-Brown}]{eggensperger-bayesopt13}
Eggensperger K, Feurer M, Hutter F, Bergstra J, Snoek J, Hoos H, Leyton-Brown K
  (2013) Towards an empirical foundation for assessing {Bayesian} optimization
  of hyperparameters. In: NIPS Workshop on {B}ayesian Optimization in Theory
  and Practice (BayesOpt'13)

\bibitem[{Eggensperger et~al(2015)Eggensperger, Hutter, Hoos, and
  Leyton-Brown}]{eggensperger-aaai15}
Eggensperger K, Hutter F, Hoos H, Leyton-Brown K (2015) Efficient benchmarking
  of hyperparameter optimizers via surrogates. In:  \cite{aaai15}, pp
  1114--1120

\bibitem[{Fawcett and Hoos(2016)}]{fawcett-heu16a}
Fawcett C, Hoos H (2016) Analysing differences between algorithm configurations
  through ablation. Journal of Heuristics 22(4):431--458

\bibitem[{Fawcett et~al(2014)Fawcett, Vallati, Hutter, Hoffmann, Hoos, and
  Leyton{-}Brown}]{fawcett-icaps14a}
Fawcett C, Vallati M, Hutter F, Hoffmann J, Hoos H, Leyton{-}Brown K (2014)
  Improved features for runtime prediction of domain-independent planners. In:
  Chien S, Minh D, Fern A, Ruml W (eds) Proceedings of the Twenty-Fourth
  International Conference on Automated Planning and Scheduling (ICAPS-14),
  {AAAI}

\bibitem[{Feurer et~al(2015{\natexlab{a}})Feurer, Klein, Eggensperger,
  Springenberg, Blum, and Hutter}]{feurer-nips2015a}
Feurer M, Klein A, Eggensperger K, Springenberg JT, Blum M, Hutter F
  (2015{\natexlab{a}}) Efficient and robust automated machine learning. In:
  Cortes C, Lawrence N, Lee D, Sugiyama M, Garnett R (eds) Proceedings of the
  29th International Conference on Advances in Neural Information Processing
  Systems (NIPS'15), pp 2962--2970

\bibitem[{Feurer et~al(2015{\natexlab{b}})Feurer, Springenberg, and
  Hutter}]{feurer-aaai15a}
Feurer M, Springenberg T, Hutter F (2015{\natexlab{b}}) Initializing {B}ayesian
  hyperparameter optimization via meta-learning. In:  \cite{aaai15}, pp
  1128--1135

\bibitem[{Gama and Brazdil(1995)}]{gama_pai95}
Gama J, Brazdil P (1995) Characterization of classification algorithms. In:
  Proceedings of the 7th {Portuguese} {Conference} on {Artificial}
  {Intelligence}, Springer, pp 189--200, to read

\bibitem[{Gebser et~al(2011)Gebser, Kaminski, Kaufmann, Schaub, Schneider, and
  Ziller}]{gebser-lpnmr11a}
Gebser M, Kaminski R, Kaufmann B, Schaub T, Schneider M, Ziller S (2011) A
  portfolio solver for answer set programming: Preliminary report. In:
  Delgrande J, Faber W (eds) Proceedings of the Eleventh International
  Conference on Logic Programming and Nonmonotonic Reasoning (LPNMR'11),
  Springer-Verlag, Lecture Notes in Computer Science, vol 6645, pp 352--357

\bibitem[{Gebser et~al(2012)Gebser, Kaufmann, and Schaub}]{gebser-ai12a}
Gebser M, Kaufmann B, Schaub T (2012) Conflict-driven answer set solving: From
  theory to practice. Artificial Intelligence 187-188:52--89

\bibitem[{Gelbart et~al(2014)Gelbart, Snoek, and Adams}]{gelbart_uai14a}
Gelbart M, Snoek J, Adams R (2014) Bayesian optimization with unknown
  constraints. In: Zhang N, Tian J (eds) Proceedings of the 30th conference on
  Uncertainty in Artificial Intelligence (UAI'14), AUAI Press

\bibitem[{Gerevini and Serina(2002)}]{gerevini-aips02}
Gerevini A, Serina I (2002) {LPG:} {A} planner based on local search for
  planning graphs with action costs. In: Ghallab M, Hertzberg J, Traverso P
  (eds) Proceedings of the Sixth International Conference on Artificial
  Intelligence, {AAAI} Press / The {MIT} Pres, pp 13--22

\bibitem[{Gorissen et~al(2010)Gorissen, Couckuyt, Demeester, Dhaene, and
  Crombecq}]{gorissen-jmlr2010a}
Gorissen D, Couckuyt I, Demeester P, Dhaene T, Crombecq K (2010) A surrogate
  modeling and adaptive sampling toolbox for computer based design. Journal of
  Machine Learning Research 11:2051--2055

\bibitem[{Guerra et~al(2008)Guerra, Prud\^encio, and
  Ludermir}]{guerra-icann08a}
Guerra S, Prud\^encio R, Ludermir T (2008) Predicting the performance of
  learning algorithms using support vector machines as meta-regressors. In:
  Kurkova-Pohlova V, Koutnik J (eds) International Conference on Artificial
  Neural Networks (ICANN'08), Springer-Verlag, vol~18, pp 523--532

\bibitem[{Hoos(2017)}]{hoos-emp17a}
Hoos H (2017) Empirical Algorithmics. Cambridge University Press, to appear

\bibitem[{Hoos and Stützle(2004)}]{hoos-sls04}
Hoos H, Stützle T (2004) Stochastic Local Search: Foundations \& Applications.
  Morgan Kaufmann Publishers Inc.

\bibitem[{Hoos et~al(2014)Hoos, Lindauer, and Schaub}]{hoos-tplp14a}
Hoos H, Lindauer M, Schaub T (2014) claspfolio 2: Advances in algorithm
  selection for answer set programming. Theory and Practice of Logic
  Programming 14:569--585

\bibitem[{Hutter et~al(2007)Hutter, Babi\'c, Hoos, and Hu}]{hutter-fmcad07a}
Hutter F, Babi\'c D, Hoos H, Hu A (2007) Boosting verification by automatic
  tuning of decision procedures. In: O'Conner L (ed) Formal Methods in Computer
  Aided Design (FMCAD'07), IEEE Computer Society Press, pp 27--34

\bibitem[{Hutter et~al(2009)Hutter, Hoos, Leyton-Brown, and
  St{\"u}tzle}]{hutter-jair09a}
Hutter F, Hoos H, Leyton-Brown K, St{\"u}tzle T (2009) Param{ILS}: An automatic
  algorithm configuration framework. Journal of Artificial Intelligence
  Research 36:267--306

\bibitem[{Hutter et~al(2010)Hutter, Hoos, and Leyton-Brown}]{hutter-cpaior10a}
Hutter F, Hoos H, Leyton-Brown K (2010) Automated configuration of mixed
  integer programming solvers. In: Lodi A, Milano M, Toth P (eds) Proceedings
  of the Seventh International Conference on Integration of {AI} and {OR}
  Techniques in Constraint Programming (CPAIOR'10), Springer-Verlag, Lecture
  Notes in Computer Science, vol 6140, pp 186--202

\bibitem[{Hutter et~al(2011{\natexlab{a}})Hutter, Hoos, and
  Leyton-Brown}]{hutter-bayesopt11}
Hutter F, Hoos H, Leyton-Brown K (2011{\natexlab{a}}) Bayesian optimization
  with censored response data. In: NIPS workshop on Bayesian Optimization,
  Sequential Experimental Design, and Bandits (BayesOpt'11)

\bibitem[{Hutter et~al(2011{\natexlab{b}})Hutter, Hoos, and
  Leyton-Brown}]{hutter-lion11a}
Hutter F, Hoos H, Leyton-Brown K (2011{\natexlab{b}}) Sequential model-based
  optimization for general algorithm configuration. In: Coello C (ed)
  Proceedings of the Fifth International Conference on Learning and Intelligent
  Optimization (LION'11), Springer-Verlag, Lecture Notes in Computer Science,
  vol 6683, pp 507--523

\bibitem[{Hutter et~al(2014{\natexlab{a}})Hutter, L\'{o}pez-Ib\'{a}nez,
  Fawcett, Lindauer, Hoos, Leyton-Brown, and St\"utzle}]{hutter-lion14a}
Hutter F, L\'{o}pez-Ib\'{a}nez M, Fawcett C, Lindauer M, Hoos H, Leyton-Brown
  K, St\"utzle T (2014{\natexlab{a}}) Aclib: a benchmark library for algorithm
  configuration. In: Pardalos P, Resende M (eds) Proceedings of the Eighth
  International Conference on Learning and Intelligent Optimization (LION'14),
  Springer-Verlag, Lecture Notes in Computer Science

\bibitem[{Hutter et~al(2014{\natexlab{b}})Hutter, Xu, Hoos, and
  Leyton-Brown}]{hutter-aij14a}
Hutter F, Xu L, Hoos H, Leyton-Brown K (2014{\natexlab{b}}) Algorithm runtime
  prediction: Methods and evaluation. Artificial Intelligence 206:79--111

\bibitem[{Hutter et~al(2017)Hutter, Lindauer, Balint, Bayless, Hoos, and
  Leyton-Brown}]{hutter-aij17a}
Hutter F, Lindauer M, Balint A, Bayless S, Hoos H, Leyton-Brown K (2017) The
  configurable {SAT} solver challenge ({CSSC}). Artificial Intelligence
  243:1--25

\bibitem[{Kadioglu et~al(2010)Kadioglu, Malitsky, Sellmann, and
  Tierney}]{kadioglu-ecai10}
Kadioglu S, Malitsky Y, Sellmann M, Tierney K (2010) {ISAC} - instance-specific
  algorithm configuration. In:  \cite{ecai10}, pp 751--756

\bibitem[{Koenker(2005)}]{koenker-quantilebook05}
Koenker R (2005) Quantile Regression. Cambridge University Press

\bibitem[{Kotthoff(2014)}]{kotthoff-aim14a}
Kotthoff L (2014) Algorithm selection for combinatorial search problems: A
  survey. AI Magazine pp 48--60

\bibitem[{Krizhevsky et~al(2012)Krizhevsky, Sutskever, and
  Hinton}]{krizhevsky-nips12}
Krizhevsky A, Sutskever I, Hinton G (2012) {ImageNet} classification with deep
  convolutional neural networks. In:  \cite{nips12}, pp 1097--1105

\bibitem[{Köpf et~al(2000)Köpf, Taylor, and Keller}]{kopf_pkdd00}
Köpf C, Taylor C, Keller J (2000) Meta-analysis: From data characterisation
  for meta-learning to meta-regression. In: Proceedings of the {PKDD}-00
  Workshop on Data Mining, Decision Support,Meta-Learning and ILP

\bibitem[{Lang et~al(2015)Lang, Kotthaus, Marwedel, Weihs, Rahnenführer, and
  Bischl}]{lang-jscs15a}
Lang M, Kotthaus H, Marwedel P, Weihs C, Rahnenführer J, Bischl B (2015)
  Automatic model selection for high-dimensional survival analysis. Journal of
  Statistical Computation and Simulation 85:62--76

\bibitem[{Leite and Brazdil(2010)}]{leite-ecai10a}
Leite R, Brazdil P (2010) Active testing strategy to predict the best
  classification algorithm via sampling and metalearning. In:  \cite{ecai10},
  pp 309--314

\bibitem[{Leite et~al(2013)Leite, Brazdil, and Vanschoren}]{leite2012}
Leite R, Brazdil P, Vanschoren J (2013) Selecting classification algorithms
  with active testing. In: Perner P (ed) Machine Learning and Data Mining in
  Pattern Recognition, Springer-Verlag, Lecture Notes in Computer Science, pp
  117--131

\bibitem[{Leyton{-}Brown et~al(2000)Leyton{-}Brown, Pearson, and
  Shoham}]{leyton-brown-ec00a}
Leyton{-}Brown K, Pearson M, Shoham Y (2000) Towards a universal test suite for
  combinatorial auction algorithms. In: Proceedings of the International
  Conference on Economics and Computation, pp 66--76

\bibitem[{Leyton{-}Brown et~al(2009)Leyton{-}Brown, Nudelman, and
  Shoham}]{leyton-brown-acm09}
Leyton{-}Brown K, Nudelman E, Shoham Y (2009) Empirical hardness models:
  Methodology and a case study on combinatorial auctions. Journal of the ACM
  56(4)

\bibitem[{Lierler and Sch{\"{u}}ller(2012)}]{lierler-lifschitz12a}
Lierler Y, Sch{\"{u}}ller P (2012) Parsing combinatory categorial grammar via
  planning in answer set programming. Springer-Verlag, Lecture Notes in
  Computer Science, vol 7265, pp 436--453

\bibitem[{Lindauer et~al(2015)Lindauer, Hoos, Hutter, and
  Schaub}]{lindauer-jair15a}
Lindauer M, Hoos H, Hutter F, Schaub T (2015) Autofolio: An automatically
  configured algorithm selector. Journal of Artificial Intelligence Research
  53:745--778

\bibitem[{Long and Fox(2003)}]{long-jair03}
Long D, Fox M (2003) The 3rd international planning competition: Results and
  analysis. Journal of Artificial Intelligence Research {(JAIR)} 20:1--59

\bibitem[{L{\'o}pez-Ib{\'a}{\~n}ez et~al(2011)L{\'o}pez-Ib{\'a}{\~n}ez,
  Dubois-Lacoste, St{\"u}tzle, and Birattari}]{lopez-ibanez-tech11a}
L{\'o}pez-Ib{\'a}{\~n}ez M, Dubois-Lacoste J, St{\"u}tzle T, Birattari M (2011)
  The irace package, iterated race for automatic algorithm configuration. Tech.
  rep., IRIDIA, Universit{\'e} Libre de Bruxelles, Belgium,
  \urlprefix\url{http://iridia.ulb.ac.be/IridiaTrSeries/IridiaTr2011-004.pdf}

\bibitem[{L{\'{o}}pez{-}Ib{\'{a}}{\~{n}}ez
  et~al(2016)L{\'{o}}pez{-}Ib{\'{a}}{\~{n}}ez, Dubois-Lacoste, Caceres,
  Birattari, and St{\"{u}}tzle}]{lopez-ibanez-orp16}
L{\'{o}}pez{-}Ib{\'{a}}{\~{n}}ez M, Dubois-Lacoste J, Caceres LP, Birattari M,
  St{\"{u}}tzle T (2016) The irace package: Iterated racing for automatic
  algorithm configuration. Operations Research Perspectives 3:43--58

\bibitem[{Loreggia et~al(2016)Loreggia, Malitsky, Samulowitz, and
  Saraswat}]{loreggia-aaai16}
Loreggia A, Malitsky Y, Samulowitz H, Saraswat V (2016) Deep learning for
  algorithm portfolios. In: Schuurmans D, Wellman M (eds) Proceedings of the
  Thirtieth National Conference on Artificial Intelligence (AAAI'16), AAAI
  Press, pp 1280--1286

\bibitem[{Malitsky et~al(2013)Malitsky, Sabharwal, Samulowitz, and
  Sellmann}]{malitsky-ijcai13a}
Malitsky Y, Sabharwal A, Samulowitz H, Sellmann M (2013) Algorithm portfolios
  based on cost-sensitive hierarchical clustering. In: Rossi F (ed) Proceedings
  of the 23rd International Joint Conference on Artificial Intelligence
  (IJCAI'13), pp 608--614

\bibitem[{Manthey and Lindauer(2016)}]{manthey-sat16a}
Manthey N, Lindauer M (2016) Spybug: Automated bug detection in the
  configuration space of {SAT} solvers. In: Proceedings of the International
  Conference on Theory and Applications of Satisfiability Testing (SAT), pp
  554--561

\bibitem[{Manthey and Steinke(2014)}]{manthey-sat14r}
Manthey N, Steinke P (2014) Too many rooks. In:  \cite{satchallenge14}, pp
  97--98

\bibitem[{Maratea et~al(2014)Maratea, Pulina, and Ricca}]{maratea-tplp13a}
Maratea M, Pulina L, Ricca F (2014) A multi-engine approach to answer-set
  programming. Theory and Practice of Logic Programming 14:841--868

\bibitem[{Meinshausen(2006)}]{meinshausen-jmlr06a}
Meinshausen N (2006) Quantile regression forests. Journal of Machine Learning
  Research 7:983--999

\bibitem[{Neal(1995)}]{neal-phd95}
Neal R (1995) Bayesian learning for neural networks. PhD thesis, University of
  Toronto, Toronto, Canada

\bibitem[{Nudelman et~al(2003)Nudelman, Leyton-Brown, Andrew, Gomes, McFadden,
  Selman, and Shoham}]{nudelman-satcomp03}
Nudelman E, Leyton-Brown K, Andrew G, Gomes C, McFadden J, Selman B, Shoham Y
  (2003) {Satzilla} 0.9, solver description, {International SAT Competition}

\bibitem[{Nudelman et~al(2004)Nudelman, Leyton-Brown, Devkar, Shoham, and
  Hoos}]{nudelman-cp04}
Nudelman E, Leyton-Brown K, Devkar A, Shoham Y, Hoos H (2004) Understanding
  random {SAT}: Beyond the clauses-to-variables ratio. In: International
  Conference on Principles and Practice of Constraint Programming (CP'04), pp
  438--452

\bibitem[{Oh(2014)}]{minisat_hack}
Oh C (2014) Minisat hack 999ed, minisat hack 1430ed and swdia5by. In:
  \cite{satchallenge14}, p~46

\bibitem[{Penberthy and Weld(1994)}]{penberthy-aaai94}
Penberthy J, Weld D (1994) Temporal planning with continuous change. In:
  Hayes{-}Roth B, Korf R (eds) Proceedings of the 12th National Conference on
  Artificial Intelligence, {AAAI} Press / The {MIT} Press, pp 1010--1015

\bibitem[{Rasmussen and Williams(2006)}]{rasmussen-book06a}
Rasmussen C, Williams C (2006) Gaussian Processes for Machine Learning. The MIT
  Press

\bibitem[{Reif et~al(2014)Reif, Shafait, Goldstein, Breuel, and
  Dengel}]{reif-ppa14a}
Reif M, Shafait F, Goldstein M, Breuel T, Dengel A (2014) Automatic classifier
  selection for non-experts. Pattern Analysis and Applications 17(1):83--96

\bibitem[{Rice(1976)}]{rice76a}
Rice J (1976) The algorithm selection problem. Advances in Computers 15:65--118

\bibitem[{Sacks et~al(1989)Sacks, Welch, Welch, and Wynn}]{sacks-ss89a}
Sacks J, Welch W, Welch T, Wynn H (1989) Design and analysis of computer
  experiments. Statistical Science 4(4):409--423

\bibitem[{Santner et~al(2003)Santner, Williams, and Notz}]{santner-2003a}
Santner T, Williams B, Notz W (2003) The design and analysis of computer
  experiments. Springer

\bibitem[{Sarkar et~al(2015)Sarkar, Guo, Siegmund, Apel, and
  Czarnecki}]{sarkar-kbse15a}
Sarkar A, Guo J, Siegmund N, Apel S, Czarnecki K (2015) Cost-efficient sampling
  for performance prediction of configurable systems. In: Cohen M, Grunske L,
  Whalen M (eds) 30th {IEEE/ACM} International Conference on Automated Software
  Engineering, {IEEE}, pp 342--352

\bibitem[{Schilling et~al(2015)Schilling, Wistuba, Drumond, and
  Schmidt-Thieme}]{schilling_kdd15}
Schilling N, Wistuba M, Drumond L, Schmidt-Thieme L (2015) Hyperparameter
  optimization with factorized multilayer perceptrons. In: Machine Learning and
  Knowledge Discovery in Databases, Springer, pp 87--103

\bibitem[{Schmee and Hahn(1979)}]{schmee-techno79}
Schmee J, Hahn G (1979) A simple method for regression analysis with censored
  data. Technometrics 21:417--432

\bibitem[{Schneider and Hoos(2012)}]{hoos-lion12a}
Schneider M, Hoos H (2012) Quantifying homogeneity of instance sets for
  algorithm configuration. In: Hamadi Y, Schoenauer M (eds) Proceedings of the
  Sixth International Conference on Learning and Intelligent Optimization
  (LION'12), Springer-Verlag, Lecture Notes in Computer Science, vol 7219, pp
  190--204

\bibitem[{Shahriari et~al(2016)Shahriari, Swersky, Wang, Adams, and
  de~Freitas}]{shahriari-ieee16a}
Shahriari B, Swersky K, Wang Z, Adams R, de~Freitas N (2016) Taking the human
  out of the loop: {A} review of {B}ayesian optimization. Proceedings of the
  {IEEE} 104(1):148--175

\bibitem[{Silverthorn et~al(2012)Silverthorn, Lierler, and
  Schneider}]{silverthorn-iclp12a}
Silverthorn B, Lierler Y, Schneider M (2012) Surviving solver sensitivity: An
  {ASP} practitioner's guide. In: Dovier A, {Santos Costa} V (eds) Technical
  Communications of the Twenty-eighth International Conference on Logic
  Programming (ICLP'12), Leibniz International Proceedings in Informatics
  (LIPIcs), vol~17, pp 164--175

\bibitem[{Snoek et~al(2012)Snoek, Larochelle, and Adams}]{snoek-nips12a}
Snoek J, Larochelle H, Adams RP (2012) Practical {B}ayesian optimization of
  machine learning algorithms. In:  \cite{nips12}, pp 2960--2968

\bibitem[{Snoek et~al(2015)Snoek, Rippel, Swersky, Kiros, Satish, Sundaram,
  Patwary, Prabhat, and Adams}]{snoek-icml15a}
Snoek J, Rippel O, Swersky K, Kiros R, Satish N, Sundaram N, Patwary M,
  Prabhat, Adams R (2015) Scalable {B}ayesian optimization using deep neural
  networks. In:  \cite{icml15}, pp 2171--2180

\bibitem[{Soares and Brazdil(2004)}]{soares-mlj04a}
Soares C, Brazdil P (2004) A meta-learning method to select the kernel width in
  support vector regression. Machine Learning Journal 54:195--209

\bibitem[{Spearman(1904)}]{spearman-ajp04}
Spearman C (1904) The proof and measurement of association between two things.
  American Journal of Psychology 15:71--101

\bibitem[{Springenberg et~al(2016)Springenberg, Klein, Falkner, and
  Hutter}]{springenberg-nips16a}
Springenberg J, Klein A, Falkner S, Hutter F (2016) Bayesian optimization with
  robust {Bayesian} neural networks. In: Proceedings of the international
  conference on Advances in Neural Information Processing Systems (NIPS'16)

\bibitem[{Takeuchi et~al(2006)Takeuchi, Le, Sears, and
  Smola}]{takeuchi-jmlr06a}
Takeuchi I, Le Q, Sears T, Smola A (2006) Nonparametric quantile estimation.
  Journal of Machine Learning Research 7:1231--1264

\bibitem[{Thornton et~al(2013)Thornton, Hutter, Hoos, and
  Leyton-Brown}]{thornton-kdd13a}
Thornton C, Hutter F, Hoos H, Leyton-Brown K (2013) {A}uto-{WEKA}: combined
  selection and hyperparameter optimization of classification algorithms. In:
  Dhillon I, Koren Y, Ghani R, Senator T, Bradley P, Parekh R, He J, Grossman
  R, Uthurusamy R (eds) The 19th ACM SIGKDD International Conference on
  Knowledge Discovery and Data Mining (KDD'13), ACM Press, pp 847--855

\bibitem[{Vallati et~al(2013)Vallati, Fawcett, Gerevini, Hoos, and
  Saetti}]{vallati-socs13a}
Vallati M, Fawcett C, Gerevini A, Hoos H, Saetti A (2013) Automatic generation
  of efficient domain-optimized planners from generic parametrized planners.
  In: Helmert M, R{\"{o}}ger G (eds) Proceedings of the Sixth Annual Symposium
  on Combinatorial Search (SOCS'14), {AAAI} Press

\bibitem[{Wistuba et~al(2015)Wistuba, Schilling, and
  Schmidt-Thieme}]{wistuba_dsaa15}
Wistuba M, Schilling N, Schmidt-Thieme L (2015) Learning hyperparameter
  optimization initializations. In: Proceedings of the International Conference
  on Data Science and Advanced Analytics (DSAA), IEEE, pp 1--10

\bibitem[{Xu et~al(2008)Xu, Hutter, Hoos, and Leyton-Brown}]{xu-jair08a}
Xu L, Hutter F, Hoos H, Leyton-Brown K (2008) {SAT}zilla: Portfolio-based
  algorithm selection for {SAT}. Journal of Artificial Intelligence Research
  32:565--606

\bibitem[{Xu et~al(2011)Xu, Hutter, Hoos, and Leyton-Brown}]{xu-rcra11a}
Xu L, Hutter F, Hoos H, Leyton-Brown K (2011) {Hydra-MIP}: Automated algorithm
  configuration and selection for mixed integer programming. In: RCRA workshop
  on Experimental Evaluation of Algorithms for Solving Problems with
  Combinatorial Explosion at the International Joint Conference on Artificial
  Intelligence (IJCAI)

\bibitem[{Yang and Wooldridge(2015)}]{ijcai15}
Yang Q, Wooldridge M (eds) (2015) Proceedings of the 25th International Joint
  Conference on Artificial Intelligence (IJCAI'15)

\end{thebibliography}

\appendix

\section{Benchmark Descriptions}\label{app:descr}

\begin{description}
	\item[\regions] is a MIP benchmark based on the well-known IBM ILOG \cplex{} solver, applied to MIP-encoded instances of the combinatorial auction winner determination problem~\citep{leyton-brown-ec00a}. The MIP instance features used in this scenario include static~\citep{leyton-brown-acm09,kadioglu-ecai10,hutter-aij14a} and probing features~\citep{xu-rcra11a}. Even though \cplex{} has $74$ parameters, its performance can be predicted quite accurately~\citep{hutter-aij14a}.
	\item[\rcw] also uses \cplex{}, in this case to solve MIP-encoded problems from computational sustainability that model habitat preservation for endangered red-cockaded woodpeckers \citep{ahmadizadeh-cp10a,xu-rcra11a}. The configuration space and the instance features are the same as in \regions; however, \cplex{} exhibits a much larger range of performance values across \textit{RCW2} instances.
	\item[\rooks] is a benchmark from the 2014 Configurable SAT Solver Challenge (CSSC'14; \citealp{hutter-aij17a}) and is based on the SAT (and ASP) solver \clasp{}~\citep{gebser-ai12a} applied to so-called ``rooks'' instances---a variant of the $n$-queens problem with additional constraints \citep{manthey-sat14r}. We use the instance features generated by the well-known algorithm selector \textit{Satzilla}~\citep{nudelman-satcomp03,xu-jair08a,hutter-aij14a} for this and all other SAT scenarios.
	This AC benchmark is distinguished by \clasp{}'s highly structured configuration space, which contains a large number of conditional parameters.
	\item[\cf] is also a benchmark from CSSC'14; it is based on the state-of-the-art SAT solver \lingeling{}~\citep{biere-tech13a} applied to circuit-based fuzz testing instances~\citep{brummayer-sat10a}.
	With $322$ parameters, \lingeling{} has the largest configuration space of any target algorithm considered in our experiments (and also one of the largest of any SAT solver we are aware of). This gives rise to a particularly challenging AC benchmark, because many parameters range from 0 to the maximal 32-bit integer and offer more scope for reductions than improvements in performance.  
	\item[\probsatseven] is another benchmark from CSSC'14; it is based on one of the state-of-the-art local search SAT solvers, \probsat{}~\citep{probSAT} applied to 7SAT random instances. With only $9$ parameters, the configuration space is quite small. 
	\item[\minisatk] is our last benchmark from the CSSC'14; it is based on a modification of the well-known \textit{MiniSAT} solver~\citep{een-sat03}, called \minisathack{}~\citep{minisat_hack} on 3SAT random instances. The $10$ categorical parameters give raise to a configuration space of $800\ 000$ parameter configurations, making it the smallest configuration space we consider. 
	\item[\satellite] was introduced in the context of parameter importance analysis with ablation~\citep{fawcett-heu16a}. It is based on the AI planning system \lpg{}~\citep{gerevini-aips02}, which exposes $67$ parameters. In this case we study \emph{satellite} instances: planning problems arising in the control and observation scheduling of orbital satellites~\citep{long-jair03}. The instance features are a combination of native planning features and features derived by translating planning instances into SAT~\citep{fawcett-icaps14a}. 
	\item[\zeno] uses the same target algorithm, \lpg{}, as \satellite{}, in combination with instances from the \emph{zenotravel} planning domain~\citep{penberthy-aaai94}, which arise in a 
version of route planning. The default configuration of \lpg{} achieves worse performance than on \satellite{}; nevertheless, after configuration, the instances from this benchmark turn out to be easier for \lpg{}.
	\item[\ws] is based on the dual-purpose SAT/ASP solver \clasp{} applied to ASP (rather than SAT) instances. \clasp{} has a richer configuration space in the ASP domain (35 additional parameters, including 12 conditional ones); to compensate, we set the configuration budget twice as high as for \rooks{}. 
The ASP problem instances encode optimizing join order in database systems~\citep{lierler-lifschitz12a}.
To generate instance features, we used the same feature extractor, \emph{claspre}, as the state-of-the-art ASP algorithm selector \textit{claspfolio}~\citep{hoos-tplp14a}.
  \item[\svmmnist] is based on a support vector machine~\citep{cortes-mlj95} (using the \textit{libsvm} implementation via \textit{scikit-learn}) applied to the well-known MNIST data set. We optimized $6$ hyperparameters of the SVM, including the kernel (rbf, polynomial or sigmoid) and its dependent hyperparameters.
  \item[\xgboostcovertype] is based on xgboost~\citep{chen-kdd16} applied to the covertype data set. We optimized $11$ mostly continuous hyperparameters.
\end{description}

\end{document}